\documentclass[11pt]{article}
\usepackage{xcolor}
\definecolor{headergray}{RGB}{200,200,200}

\usepackage[final]{acl}

\usepackage{times}
\usepackage{latexsym}

\usepackage[T1]{fontenc}
\usepackage{amsmath}
\usepackage{booktabs}
\usepackage{enumitem}
\usepackage{pifont}
\usepackage{multirow}
\usepackage{makecell}
\usepackage{adjustbox}
\usepackage{tikz}
\usepackage{pgfplots}
\usepgfplotslibrary{groupplots}
\usepackage{pgfplotstable}
\usepackage{float}
\usepackage[table]{xcolor}
\usepackage{colortbl}
\usepackage{amssymb}
\usepackage{hyperref}
\usepackage[utf8]{inputenc}

\usepackage{microtype}

\usepackage{inconsolata}

\usepackage{graphicx}

\usepackage[most]{tcolorbox}
\usepackage{xcolor}
\usepackage{graphicx}
\usepackage{array}

\definecolor{casebg}{RGB}{248,248,248}
\definecolor{caseborder}{RGB}{210,210,210}
\definecolor{tokenblue}{RGB}{30,90,180}

\newtcbox{\promptbox}{
  on line,
  colback=blue!4,
  colframe=blue!35,
  boxrule=0.4pt,
  arc=1pt,
  left=2pt,
  right=2pt,
  top=1pt,
  bottom=1pt,
  fontupper=\ttfamily\footnotesize
}

\newtcolorbox{casebox}[1]{
  colback=casebg,
  colframe=caseborder,
  boxrule=0.6pt,
  arc=2pt,
  left=6pt,
  right=6pt,
  top=6pt,
  bottom=6pt,
  title=\textbf{#1},
  coltitle=black,
  colbacktitle=gray!12,
  fonttitle=\small,
}

\newcommand{\casefield}[1]{\vspace{3pt}\noindent\textbf{\small #1}\vspace{2pt}}
\newcommand{\casetext}[1]{\noindent{\footnotesize #1}}
\newcommand{\answerrow}[2]{
\vspace{4pt}
\noindent
\begin{tabular}{@{}p{0.48\linewidth}p{0.48\linewidth}@{}}
\textbf{\small OMG-VLM Output} & \textbf{\small Ground Truth} \\
\textit{\footnotesize #1} & \textit{\footnotesize #2}
\end{tabular}
}

\title{One Model, Many Graphs: Learning over Attributed Graphs across Heterogeneous Modalities with Vision-Language Models}

\author{
  \textbf{Jiayi Yang\textsuperscript{1*}},
  \textbf{Yifang Chen\textsuperscript{1*}},
  \textbf{Yuanfu Sun\textsuperscript{1,2}},
  \textbf{Jiajin Liu\textsuperscript{1,2}},
  \textbf{Qiaoyu Tan\textsuperscript{1\ensuremath{\dagger}}}
  \\
  \textsuperscript{1}New York University Shanghai,
  \textsuperscript{2}New York University
  \\
  \texttt{\{jy4656,yc6990,qiaoyu.tan\}@nyu.edu}
}

\begin{document}
\maketitle
\begingroup
\renewcommand{\thefootnote}{\fnsymbol{footnote}}
\footnotetext[1]{Equal contribution.}
\footnotetext[2]{Corresponding author.}
\endgroup
\begin{abstract}
Vision-language models (VLMs) provide a unified representation space for textual and visual information, yet their potential as general-purpose backbones for graph-structured data remains largely unexplored. In practice, attributed graphs exhibit substantial modality heterogeneity: some graphs contain only textual node attributes, others only visual attributes, while still others provide both. Existing graph learning approaches are typically designed for fixed modality schemas, requiring separate models for different settings and limiting scalability and cross-graph generalization. To bridge this gap, we present \textbf{OMG-VLM} (\textit{One Model, Many Graphs with Vision-Language Models}), a unified framework for learning over attributed graphs across heterogeneous modality schemas. OMG-VLM leverages a pretrained VLM as a shared backbone and introduces structure-aware graph adapters that integrate neighborhood information while remaining compatible with the VLM's native embedding space. This design enables effective learning over text-attributed, image-attributed, and multimodal-attributed graphs within a single model. Extensive experiments across diverse domains show that OMG-VLM consistently outperforms state-of-the-art GNN- and LLM-based baselines on attributed graph learning tasks such as node classification and link prediction, while exhibiting strong generalization to unseen graphs and varying modality schemas. The source code is available at \url{https://github.com/Jo-eyang/OMG-VLM}.
\end{abstract}

\section{Introduction}

\begin{figure}[t]
  \centering
  \includegraphics[width=\columnwidth]{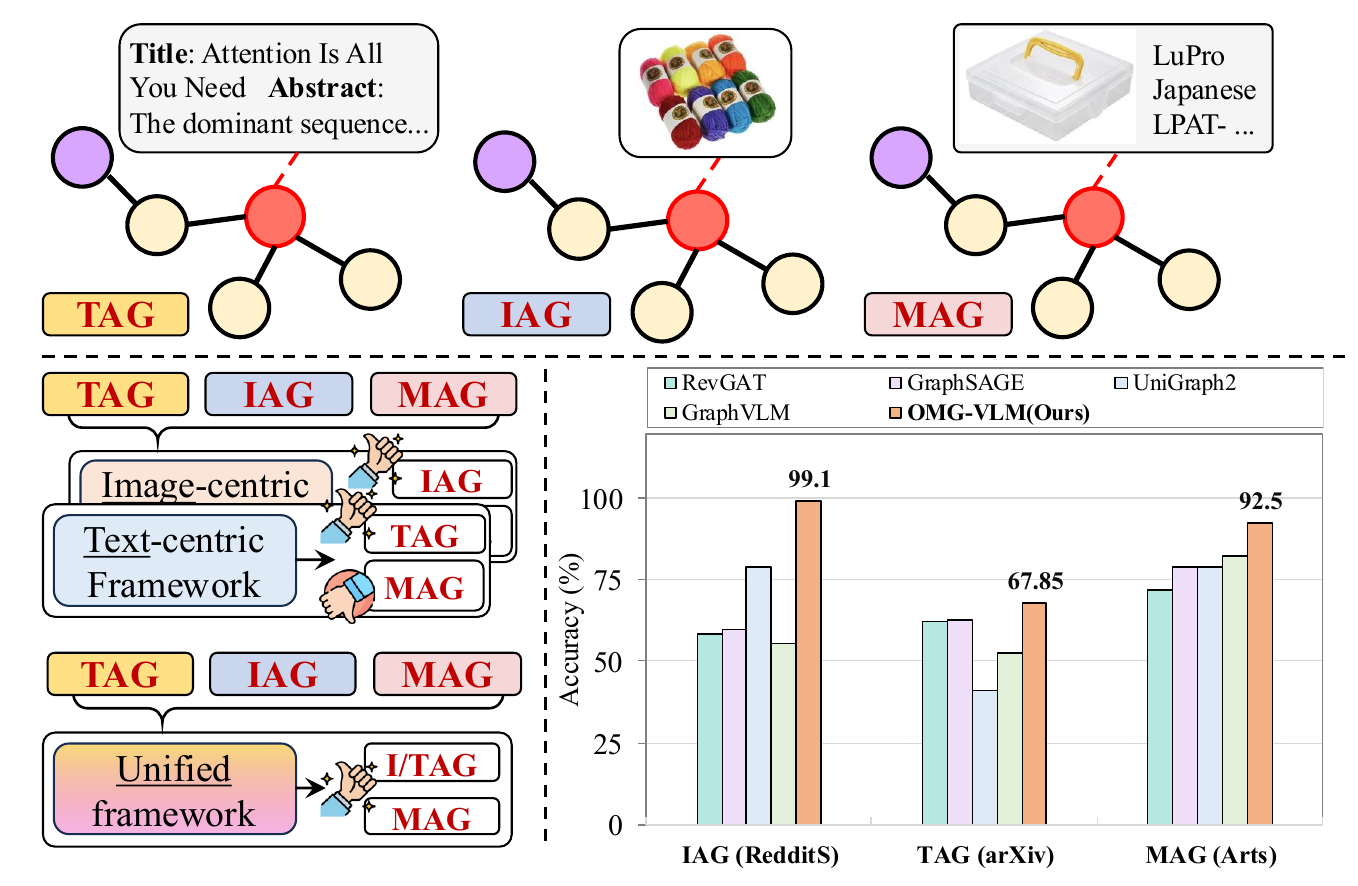}
  \caption{Heterogeneous modalities over attributed graphs. Prior methods target TAGs or MAGs, while OMG-VLM handles heterogeneous modality schemas. Bars compare OMG-VLM with representative baselines on IAG (RedditS), TAG (arXiv), and MAG (Arts).}
  \label{fig:graphs} 
  \vspace{-1.2em}
\end{figure}

Attributed graph learning is a fundamental problem in data mining and machine learning, with applications in social networks, e-commerce platforms, citation graphs, and knowledge-centric systems \cite{kipf_semi-supervised_2017, hamilton_inductive_2017, hu_open_2020, velickovic_graph_2018}. In these settings, node attributes provide semantic signals beyond graph topology, enabling models to reason jointly over structure and content. However, attribute modalities vary substantially across domains: some graphs contain textual attributes, such as titles or abstracts in citation networks; others contain visual attributes, such as product or user images; and some combine multiple modalities \cite{wei_mmgcn_2019,wang_dualgnn_2023,he_ups_2016,ning_graph4mm_2025}. This modality heterogeneity reflects real-world data collection processes, where available attributes depend on domain-specific constraints. Developing models that can operate across such heterogeneous attributed graphs is therefore an important yet underexplored challenge.

Existing graph learning methods are largely built around fixed modality assumptions. GNNs typically operate on predefined feature vectors and require modality-specific encoders to convert raw attributes into numeric representations \cite{kipf_semi-supervised_2017, velickovic_graph_2018,yan_comprehensive_2023,he_harnessing_2024,zolnai2024stage}. This often leads to separate architectures or preprocessing pipelines for text-attributed, image-attributed, and multimodal graphs, limiting reuse and cross-graph generalization. Recent LLM-based graph methods provide more flexibility for text-attributed graphs by incorporating neighborhood information through prompting or in-context learning \cite{lv_graphprompter_2025,tang_graphgpt_2024,chen_llaga_2024,he_unigraph_2025}. However, they remain text-centric and typically rely on external vision encoders for visual attributes~\cite{fan_mlaga_2026,yan2025graph,he_unigraph_2025}, which limits their ability to support image-only or heterogeneous multimodal graphs within a unified model.

To address this gap, we study \textbf{attributed graph learning under modality heterogeneity through the lens of VLMs}. Pretrained VLMs provide a shared representation space for textual and visual information \cite{bai_qwen-vl_2023,liu_visual_2023}, making them a natural foundation for unified learning across text-attributed, image-attributed, and multimodal-attributed graphs. Instead of relying on separate modality-specific encoders, we investigate whether a single VLM-based model can serve as a shared backbone across heterogeneous attributed graphs. Such a model can reduce the need for separate architectures, support transfer across graphs with different attribute schemas, and improve scalability in real-world settings where modality availability varies across domains, as motivated in Figure~\ref{fig:graphs}.

However, applying VLMs to heterogeneous attributed graphs remains challenging. First, graph neighborhoods introduce variable-sized and structure-dependent context that does not naturally fit the sequential input format of VLMs, making naïve neighborhood injection inefficient and prone to noise \cite{chen_tokenized_2023}. Second, graph-derived information must be integrated in a way that is compatible with the VLM's native textual and visual processing pipeline. External encoders or mismatched representation spaces may introduce additional gaps, making unified multimodal graph learning more difficult \cite{shu_large_2025}.

We propose \textbf{OMG-VLM} (\textit{One Model, Many Graphs with Vision-Language Models}), a unified framework for attributed graph learning across heterogeneous modality schemas. OMG-VLM uses a pretrained VLM as a shared backbone and introduces structure-aware graph adapters to inject neighborhood information while remaining native to the VLM's embedding space. It further employs modality-specific but backbone-native mechanisms to incorporate textual and visual neighborhoods, together with a co-optimization strategy that enables effective adaptation to graph-structured inputs. Our \textbf{contributions} are summarized as follows:

\begin{itemize}[leftmargin=*, topsep=0mm]
\item We study \textbf{unified attributed graph learning under heterogeneous modality schemas}, where a single model operates across text-attributed, image-attributed, and multimodal-attributed graphs. This setting reflects realistic deployment scenarios in which different graphs expose different attribute modalities, while existing methods typically rely on modality-specific designs, only handling graphs with a specific modality schema. To the best of our knowledge, this is the first work to systematically study this problem from a VLM perspective.
\item We propose \textbf{OMG-VLM}, a unified VLM-based framework for attributed graph learning across heterogeneous modality schemas. OMG-VLM introduces structure-aware graph adapters that incorporate neighborhood information compatible with the VLM's native representation space, enabling end-to-end learning across graphs with different modality configurations.
\item Through extensive experiments on node classification and link prediction across diverse domains and modality schemas, we show that OMG-VLM consistently \textbf{outperforms state-of-the-art GNN- and LLM-based baselines}. It also demonstrates strong generalization capabilities across various domains and modality schemas, supporting the effectiveness of VLMs as unified backbones for attributed graph learning.
\end{itemize}

\begin{figure*}[ht]
  \centering
  \includegraphics[width=\textwidth]{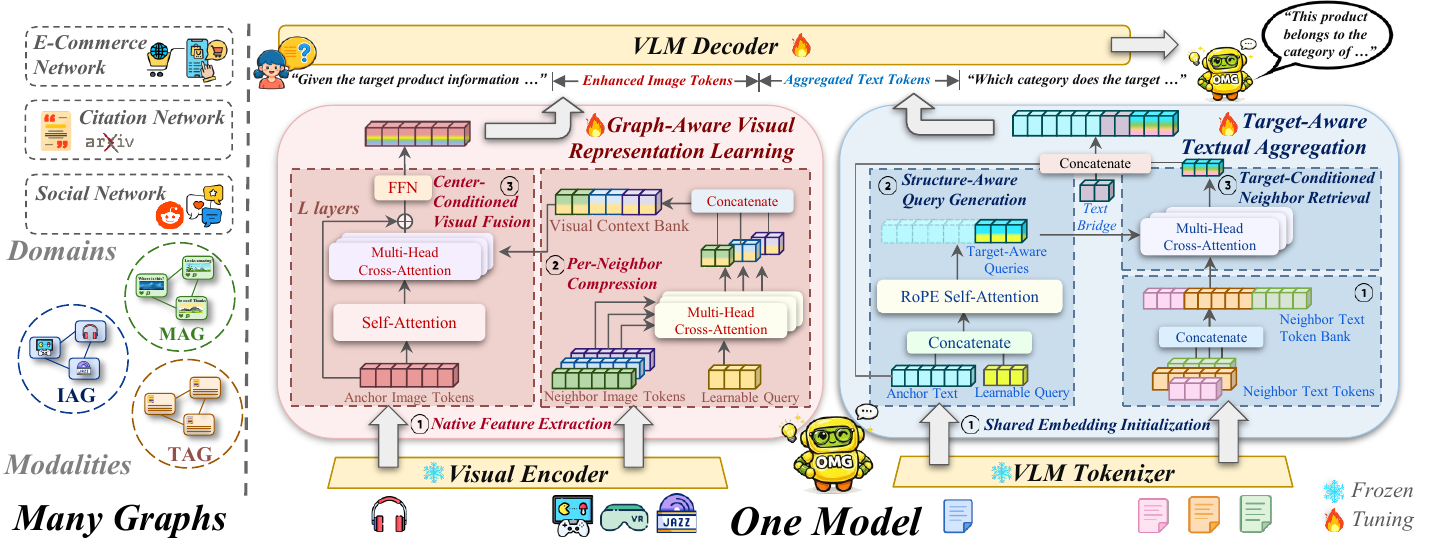}
  \caption{Overview of the OMG-VLM framework. OMG-VLM uses a pretrained VLM as a shared backbone for reasoning over attributed graphs with heterogeneous modality schemas. Textual and visual neighborhoods are incorporated via \textit{Target-Aware Textual Aggregation} and \textit{Graph-Aware Visual Representation Learning}, enabling a single model to handle TAGs, IAGs, and MAGs within a shared embedding space.}
  \label{fig:pipeline}  
\end{figure*}
\section{Related Works}
\noindent\textbf{Text-Attributed Graph (TAG) Learning.}
GNNs, including GCN \cite{kipf_semi-supervised_2017}, GraphSAGE \cite{hamilton_inductive_2017}, and RevGAT \cite{li_training_2021}, are widely used for node- and link-level prediction via neighborhood propagation. For TAGs, prior work commonly encodes node texts with pretrained language models and then applies graph propagation~\cite{he_unigraph_2025,he_harnessing_2024,10.1145/3655103.3655110,10.1145/3732786,yan_comprehensive_2023}. Recent LLM-based methods instead serialize graph neighborhoods and structural information into textual prompts for generative graph reasoning \cite{chen_llaga_2024,tang_graphgpt_2024,lv_graphprompter_2025,sun_graphicl_2025,zhang_graphtranslator_2024,sun2026agentgl}. Beyond prompting, hybrid approaches combine language and graph models. For example, SSTAG~\cite{liu_sstag_2025} co-distills LLM and GNN knowledge into structure-aware MLPs for scalable TAG learning. However, these methods remain text-centric and are not designed for non-textual node attributes.

\noindent\textbf{Multimodal Attributed Graph (MAG) Learning.}
MAG learning extends TAGs by incorporating both textual and visual node attributes, requiring models to jointly capture multimodal features and graph structure~\cite{yan2025graph}. Early methods are largely GNN-based and often target domain-specific applications such as recommendation \cite{wei_mmgcn_2019,wang_dualgnn_2023}. Recent work leverages multimodal encoders such as CLIP \cite{radford_learning_2021} for multimodal attributed graph reasoning~\cite{zhu2025mosaic,liu2026graphvlm,ning_graph4mm_2025,yan2025graph,sun2026mario}. UniGraph2~\cite{he_unigraph2_2025} converts modality-specific features into unified low-dimensional node embeddings through modality alignment and graph propagation, with downstream tasks handled by separate predictors. It therefore focuses on representation-level rather than prediction-level unification. MLaGA~\cite{fan_mlaga_2026} and Graph4MM~\cite{ning_graph4mm_2025} instead integrate multimodal features through external encoders and fusion modules. Although effective for MAGs, these methods primarily target fully multimodal schemas. A single generative model operating across TAGs, IAGs, and MAGs through one prediction interface remains underexplored.

\noindent\textbf{Vision-Language Models (VLMs).}
VLMs align visual and textual representations through multimodal pretraining and instruction tuning, enabling strong visual understanding and language reasoning \cite{li_blip-2_2023,liu_visual_2023,bai_qwen-vl_2023}. However, they are not designed to process graph-structured context natively. GraphVLM \cite{liu2026graphvlm} applies VLMs to graphs by serializing nodes, edges, or structural descriptions into prompts. While promising, this prompt-based linearization is constrained by input length and primarily targets fully multimodal attributed graphs.

\section{Problem Statement}
\label{sec:problem}

\noindent\textbf{Attributed Graph Learning with Heterogeneous Modality Schemas.}
We are given a collection of attributed graphs
$\{\mathcal{G}_m=(\mathcal{V}_m,\mathcal{E}_m)\}_{m=1}^{K}$,
where different graphs may follow different modality schemas. We consider three common cases:
(1) text-attributed graphs (TAGs), where nodes are associated with textual attributes;
(2) image-attributed graphs (IAGs), where nodes are associated with visual attributes; and
(3) multimodal-attributed graphs (MAGs), where nodes are associated with both textual and visual attributes.

For a node $v \in \mathcal{V}_m$, let $T_v$ and $I_v$ denote its textual and visual attributes. Either modality may be absent depending on the graph schema. We denote the available modality set of $v$ as $\mathcal{A}_v \subseteq \{T_v, I_v\}.$
Similarly, let $\mathcal{N}(v)$ denote the neighborhood of $v$. We define $\mathcal{N}_T(v)$ and $\mathcal{N}_I(v)$ as the available respective textual and visual attributes associated with neighbors in $\mathcal{N}(v)$, depending on availability.

\noindent\textbf{Learning Objective.}
We study generative reasoning on attributed graphs with heterogeneous modality schemas. Given a target node $v$, its available attributes $\mathcal{A}_v$, its modality-specific neighborhood context $\mathcal{N}_T(v)$ and $\mathcal{N}_I(v)$, and a task instruction $\mathcal{P}$, the goal is to learn a unified model that generates the target sequence $Y$ under a shared conditional generation objective:
\begin{equation}
\label{eq:loss}
\small
\mathcal{L}(\Theta)
=
\log P
\big(
Y
\mid
\mathcal{A}_v,
\mathcal{N}_T(v),
\mathcal{N}_I(v),
\mathcal{P};
\Theta
\big),
\end{equation}
where $\Theta$ denotes the model parameters. In this paper, we focus on two representative graph reasoning tasks: node classification and link prediction. For node classification, the query is a target node $v$ and $Y$ denotes its label. For link prediction, the query is a node pair $(v_i,v_j)$ and $Y$ denotes whether an edge exists between them.

\section{Methodology}
\label{sec:4}
We introduce OMG-VLM, a generative framework for reasoning over attributed graphs with heterogeneous modality schemas. We first formulate graph reasoning as generative modeling (Sec.~\ref{sec:4-1}), then present two graph adapters for incorporating neighborhood context: \textit{Target-Aware Textual Aggregation} (Sec.~\ref{sec:4-2}) and \textit{Graph-Aware Visual Representation Learning} (Sec.~\ref{sec:4-3}). Finally, we describe a unified training objective for optimizing these modules within the shared VLM backbone (Sec.~\ref{sec:4-4}).
\subsection{Overview}
\label{sec:4-1}

OMG-VLM builds graph reasoning on top of a pretrained VLM with parameters $\Theta$. The key idea is to represent heterogeneous graph context as a sequence of VLM-compatible tokens, so that attributed graphs with different modality schemas can be handled by the same generative backbone. 
Rather than introducing separate modality encoders and then aligning them with the VLM, OMG-VLM operates directly in the VLM's native multimodal embedding space. 
Textual and visual node attributes, together with their modality-specific neighborhoods, are first transformed into compact token representations by graph adapters and then concatenated with the task instruction. Unlike embedding-level unification followed by separate downstream predictors, OMG-VLM maps modality-specific neighborhoods into the VLM's native token space and decodes predictions through one generative backbone and output interface across TAGs, IAGs, and MAGs. Because the adapters produce VLM-compatible token sequences, the model preserves target-relevant textual evidence and patch-level visual information without compressing all graph context into a single generic node vector.

Given a target node $v$, OMG-VLM constructs the input sequence $\mathbf{X}_v$ as

\begin{small}
\begin{equation}
\label{eq:input_construction}
\mathbf{X}_v =
\underbrace{
\Phi_T(T_v,\mathcal{N}_T(v))
\ \Vert\
\Phi_I(I_v,\mathcal{N}_I(v))
}_{\text{graph context}}
\ \Vert\
\underbrace{\mathbf{E}_{\mathcal{P}}}_{\text{instruction}},
\end{equation}
\end{small}
where $\mathbf{E}_{\mathcal{P}}=\mathrm{Embed}(\mathcal{P})$. 
Here, $\Phi_T$ and $\Phi_I$ denote the textual and visual graph adapters, which map modality-specific node attributes and neighborhood context into compact token sequences compatible with the VLM backbone. 
If a modality is unavailable for a given graph, the corresponding adapter output is omitted. 
This design allows the same model to process graphs with heterogeneous modality schemas under a unified input format, as illustrated in Figure~\ref{fig:pipeline}.

Conditioned on $\mathbf{X}_v$, the VLM generates the target output sequence autoregressively:
\begin{equation}
\label{eq:generation}
P(Y \mid \mathcal{G}, \mathcal{P}) =
\prod_{t=1}^{|Y|}
P_{\mathrm{VLM}}
\big(
y_t \mid y_{<t}, \mathbf{X}_v; \Theta
\big).
\end{equation}

By converting graph-derived textual and visual context into native VLM token representations, OMG-VLM avoids explicit cross-modal realignment while preserving the pretrained multimodal interface of the backbone. 
The following sections describe how $\Phi_T$ and $\Phi_I$ aggregate textual and visual neighborhoods, respectively, before jointly optimizing them with the generative objective.

\subsection{Target-Aware Textual Aggregation}
\label{sec:4-2}

Textual neighbors provide useful graph context, but their relevance to the target node varies. Existing methods typically encode neighborhood text with external encoders or concatenate it into the input \cite{liu2026graphvlm, chen_llaga_2024, fan_mlaga_2026}. For VLM backbones, however, external encoders require additional cross-modal alignment, while concatenation yields noisy and inefficient inputs. OMG-VLM instead incorporates textual graph context through an adapter built on top of the VLM's own textual representations, reducing the need for explicit cross-space alignment.

We introduce \textit{Target-Aware Textual Aggregation} $\Phi_T$, which performs target-conditioned token-level retrieval over textual neighbors.

\noindent\textbf{Shared Embedding Initialization.}
The target text and textual neighbor attributes are embedded using the token embedding matrix of the pretrained VLM. Given the target text $T_v$, its token embeddings are:
\begin{equation}
\label{eq:text_target_embed}
\mathbf{E}_v^T
=
\mathrm{Embed}
\big(
\mathrm{Tokenizer}(T_v)
\big)
\in \mathbb{R}^{L_v^T \times d},
\end{equation}
where $L_v^T$ is the target text length and $d$ is the hidden dimension of the backbone VLM.

For textual neighbors $\mathcal{N}_T(v)$, we concatenate all token embeddings into a textual neighbor bank:
\begin{equation}
\label{eq:text_neighbor_bank}
\small
\mathbf{B}_v^T
=
\mathop{\Big\Vert}_{u \in \mathcal{N}_T(v)}
\mathrm{Embed}
\big(
\mathrm{Tokenizer}(T_u)
\big)
\in \mathbb{R}^{L_{\mathcal{N}}^T \times d},
\end{equation}
where $L_{\mathcal{N}}^T$ is the number of textual neighbor tokens.

\noindent\textbf{Target-Aware Query Generation.}
Since the VLM tokenizer lacks an explicit pooling token such as \texttt{[CLS]} \cite{bai_qwen-vl_2023,devlin_bert_2019}, we introduce $M_T$ learnable query tokens
$\mathbf{Q}_T \in \mathbb{R}^{M_T \times d}$ as adaptive semantic probes. We concatenate them with the target text embeddings and apply a RoPE-aware self-attention block~\cite{su_roformer_2024}:
\begin{equation}
\label{eq:text_query_generation}
\mathbf{H}_v^T
=
\mathrm{Attn}_{\mathrm{RoPE}}
\big(
\mathbf{E}_v^T
\Vert
\mathbf{Q}_T
\big).
\end{equation}
The final $M_T$ tokens are used as target-conditioned textual queries, which summarize multiple semantic aspects of the center node:
\begin{equation}
\label{eq:text_target_query}
\widehat{\mathbf{Q}}_v^T
=
\mathbf{H}_v^T[-M_T:,:]
\in \mathbb{R}^{M_T \times d}.
\end{equation}

\noindent\textbf{Target-Conditioned Neighbor Retrieval.}
Given $\widehat{\mathbf{Q}}_v^T$, we retrieve informative textual neighborhood content through multi-head cross-attention:
\begin{equation}
\label{eq:text_neighbor_retrieval}
\mathbf{Z}_v^T
=
\mathrm{MHCA}
\big(
\widehat{\mathbf{Q}}_v^T,
\mathbf{B}_v^T,
\mathbf{B}_v^T
\big)
\in \mathbb{R}^{M_T \times d}.
\end{equation}
This relevance-based retrieval preserves fine-grained neighborhood information, avoiding the uniform averaging in static pooling strategies.

The final textual adapter output is:
\begin{equation}
\label{eq:text_adapter_output}
\Phi_T(T_v,\mathcal{N}_T(v))
=
\mathbf{X}_v^T
=
\mathbf{E}_v^T
\Vert
\mathbf{E}_{\mathrm{sep}}^T
\Vert
\mathbf{Z}_v^T,
\end{equation}
where $\mathbf{E}_{\mathrm{sep}}^T$ is a delimiter embedding acting as a text bridge, e.g., ''Neighbor Text Information:''.

\subsection{Graph-Aware Visual Representation Learning}
\label{sec:4-3}

Visual neighborhoods pose a distinct challenge for graph reasoning. Each image is represented as patch tokens, so directly incorporating $K$ visual neighbors yields $KL^I$ tokens, where $L^I$ is the patch sequence length per image. Aggregating these high-dimensional, spatially redundant tokens can dilute salient cues and incur substantial computational overhead during inference.

To address this challenge, we introduce a \textit{Graph-Aware Visual Adapter} $\Phi_I$ with two stages: (i) per-neighbor visual compression and (ii) center-conditioned visual aggregation.

\noindent\textbf{Native Feature Extraction.}
We first extract visual features using the VLM's frozen visual encoder. In modern VLMs, a Vision Transformer (ViT) serves as the inherent visual encoder that maps raw images into visual token sequences~\cite{dosovitskiy_image_2021, bai_qwen-vl_2023}. For the target node $v$ and each visual neighbor $u \in \mathcal{N}_I(v)$:
\begin{equation}
\label{eq:visual_feature_extraction}
\small
\mathbf{E}_u^I
=
\mathrm{ViT}(I_u)
\in \mathbb{R}^{L^I \times d},
\quad
u \in \{v\} \cup \mathcal{N}_I(v),
\end{equation}
where $L^I$ is the number of visual tokens and $d$ is the hidden dimension of the backbone VLM. The target visual tokens $\mathbf{E}_v^I$ serve as the reference for visual neighborhood aggregation.

\noindent\textbf{Per-Neighbor Compression.}
Each visual neighbor is compressed into $M_I$ tokens using learnable visual queries
$\mathbf{Q}_I \in \mathbb{R}^{M_I \times d}$:
\begin{equation}
\label{eq:visual_neighbor_compression}
\small
\mathbf{C}_u^I
=
\mathrm{MHCA}\!\left(\mathbf{Q}_I,\mathbf{E}_u^I,\mathbf{E}_u^I\right),
\quad u \in \mathcal{N}_I(v).
\end{equation}
Here, $\mathbf{C}_u^I \in \mathbb{R}^{M_I \times d}$ and $M_I \ll L^I$, reducing the visual context from
$|\mathcal{N}_I(v)|L^I$ to $|\mathcal{N}_I(v)|M_I$ tokens, while enabling adaptive selection of informative visual regions.

The compressed visual neighbor tokens are concatenated into a visual context bank:
\begin{equation}
\label{eq:visual_neighbor_bank}
\mathbf{B}_v^I
=
\mathop{\Big\Vert}_{u \in \mathcal{N}_I(v)}
\mathbf{C}_u^I
\in
\mathbb{R}^{|\mathcal{N}_I(v)|M_I \times d}.
\end{equation}

\noindent\textbf{Center-Conditioned Visual Aggregation.}
OMG-VLM integrates visual neighborhood information by refining the target visual tokens in place. At layer $\ell$, the target visual tokens attend to the compressed visual context bank:

{
\begin{small}
\begin{equation*}
\label{eq:visual_aggregation}
\mathbf{H}_v^{I,(\ell+1)}
=
\mathrm{FFN}\!\left(
\mathrm{MHCA}\!\left(
\mathbf{H}_v^{I,(\ell)}, \mathbf{B}_v^I, \mathbf{B}_v^I
\right)
+
\mathbf{H}_v^{I,(\ell)}
\right)
\end{equation*}
\end{small}
}
with $\mathbf{H}_v^{I,(0)}=\mathbf{E}_v^I$.

After $L$ aggregation layers, the resulting graph-aware visual representation is:
\begin{equation}
\label{eq:visual_adapter_output}
\Phi_I(I_v,\mathcal{N}_I(v))=
\mathbf{X}_v^I
=
\mathbf{H}_v^{I,(L)}.
\end{equation}
The refined tokens preserve the original visual token length, allowing direct substitution into the VLM input sequence without increasing downstream sequence length.

\subsection{Co-Optimization on Attributed Graphs with Heterogeneous Modality Schemas}
\label{sec:4-4}

A key advantage of leveraging a pretrained VLM is that textual and visual inputs are mapped into a shared embedding space by the backbone tokenizer and visual encoder. This unified representation reduces the need for explicit cross-modal alignment, which is often required in LLM-based graph methods for TAGs or MAGs. OMG-VLM therefore supports joint optimization while preserving pretrained multimodal representations.

\noindent\textbf{Training Strategy.}
We jointly optimize the LoRA-adapted VLM parameters and the two graph adapters, while keeping the remaining backbone parameters frozen. This adapts the VLM to graph-structured inputs while preserving its native visual encoder and tokenizer.

Training uses a heterogeneous collection of attributed graphs, including TAGs, IAGs, and MAGs, across multiple domains and graph reasoning tasks. This exposes OMG-VLM to diverse modality schemas, encouraging structure-aware reasoning that generalizes beyond a single modality schema.

Let $\mathcal{D}$ be training set, where each example is:
\begin{equation}
\label{eq:training_example}
z
=
(v,T_v,I_v,\mathcal{N}_T(v),\mathcal{N}_I(v),\mathcal{P},Y).
\end{equation}
The input sequence $\mathbf{X}_v$ is constructed by Eq.~\ref{eq:input_construction} according to the modalities available in each example. The model maximizes:
\begin{equation}
\label{eq:train_objective}
\small
\mathcal{L}(\Theta)
=
\mathbb{E}_{z \sim \mathcal{D}}
\left[
\sum_{t=1}^{|Y|}
\log
P_{\mathrm{VLM}}
\big(
y_t
\mid
y_{<t},
\mathbf{X}_v;
\Theta
\big)
\right].
\end{equation}

All trainable components are updated jointly via backpropagation, allowing neighborhood retrieval, aggregation, and generation to co-adapt during optimization. This enables OMG-VLM to operate as a single model across attributed graphs with heterogeneous modality schemas.

\begin{table*}[t]
\centering
\scriptsize
\setlength{\tabcolsep}{2pt}
\renewcommand{\arraystretch}{1.05}

\resizebox{\textwidth}{!}{
\begin{tabular}{ll | ccccc | cccccc | c}
\toprule
\rowcolor{headergray}

\textbf{Category} & \textbf{Model}
& \multicolumn{5}{c|}{\textbf{In-Domain}} 
& \multicolumn{6}{c|}{\textbf{Transfer}} 
& \textbf{Overall} \\
\cmidrule(lr){3-7} \cmidrule(lr){8-13} \cmidrule(lr){14-14}

& 
& Arts$^{\text{\scriptsize NC}}$ 
& arXiv$^{\text{\scriptsize NC}}_{\dagger}$ 
& Movies$^{\text{\scriptsize LP}}$ 
& RedditS$^{\text{\scriptsize LP}}_{\diamond}$ 
& \textbf{Avg.}
& PubMed$^{\text{\scriptsize NC}}_{\dagger}$ 
& CDs$^{\text{\scriptsize NC}}$
& RedditM$^{\text{\scriptsize LP}}$ 
& VideoGames$^{\text{\scriptsize LP}}$ 
& Cora$^{\text{\scriptsize LP}}_{\dagger}$ 
& \textbf{Avg.} 
& \textbf{Avg.} \\

\midrule

\textbf{MLP} 
& MLP 
& 73.75 & \underline{64.59} & 59.42 & 56.25 & 63.50 
& 37.96 & 1.42 & 55.20 & 50.12 & 50.90 & 39.12 & 49.96 \\

\midrule

\textbf{GNNs \&} 
& GCN 
& 63.95 & 58.71 & 52.38 & 49.77 & 56.20 
& 20.79 & 10.79 & 51.05 & 50.61 & 50.00 & 36.65 & 45.34 \\

\textbf{GNN-based} 
& RevGAT 
& 71.89 & 62.15 & 49.54 & 49.77 & 58.34 
& 40.26 & 3.54 & 50.25 & 50.08 & 50.00 & 38.83 & 47.50 \\

& GraphSAGE 
& 74.77 & 62.78 & 52.22 & 49.87 & 59.91 
& 42.77 & 3.67 & 50.80 & 52.99 & 50.30 & 40.11 & 48.91 \\

& UniGraph2 
& 78.87 & 41.26 & \underline{91.51} & 78.85 & 72.62 
& 38.40 & 10.83 & 62.35 & 54.41 & \underline{51.10} & 43.42 & 56.40 \\

\midrule

\textbf{LLM-based}
& LLaGA 
& 88.74 & 63.70 & 53.27 & -- & 68.57 
& 1.65 & 4.14 & 51.60 & 57.98 & 0.70 & 23.21 & 40.22 \\

& GraphPrompter 
& 83.21 & 56.28 & 63.82 & -- & 67.77 
& 56.92 & 30.71 & 44.65 & \underline{70.30} & 50.00 & 50.52 & 56.99 \\

\midrule

\textbf{LLM w/} 
& MLaGA 
& \underline{89.63} & -- & 76.20 & \underline{90.10} & \underline{85.31} 
& -- & \underline{31.10} & 64.30 & 58.46 & -- & 51.29 & \underline{68.30} \\

\textbf{Ext. Enc.} 
& Graph4MM 
& 84.05 & 52.75 & 88.45 & -- & 75.08 
& 25.95 & 10.25 & \underline{72.30} & 59.40 & 14.90 & 36.56 & 51.01 \\

\midrule

\textbf{VLM-based}
& GraphVLM 
& 82.55 & 63.99 & 50.74 & 55.70 & 63.25 
& \underline{72.13} & 29.47 & 57.35 & 48.73 & 50.20 & \underline{51.58} & 56.76 \\

& \textbf{OMG-VLM} 
& \textbf{92.50} & \textbf{67.85} 
& \textbf{95.30} & \textbf{99.10} & \textbf{88.69}
& \textbf{78.55} & \textbf{32.56}
& \textbf{92.45} & \textbf{77.99} & \textbf{51.80} & \textbf{66.67} & \textbf{76.46} \\

\bottomrule
\end{tabular}
}

\caption{Main results under in-domain and transfer settings. In-domain results are evaluated on the four training graphs, while transfer results are evaluated on unseen graphs. Subscripts ${}_{\dagger}$ and ${}_{\diamond}$ denote TAG and IAG datasets, respectively; all remaining datasets are MAGs. Bold and underline indicate the best and second-best results. -- indicates that a model cannot process the dataset due to architectural constraints. Additional baseline implementation details and differences from originally reported results are discussed in Appendix~\ref{app:baseline_details}.}
\vspace{+1em}
\label{tab:main_results}
\end{table*}

\section{Experiments}
We conduct experiments to address the following four research questions (\textbf{RQs}):

\noindent\textbf{RQ1:} How does OMG-VLM compare with graph reasoning baselines on in-domain tasks?

\noindent\textbf{RQ2:} How well does OMG-VLM generalize to unseen graphs under transfer settings?

\noindent\textbf{RQ3:} How much do the textual adapter, visual adapter, and co-optimization strategy contribute?

\noindent\textbf{RQ4:} How sensitive is OMG-VLM to key hyperparameters for graph context modeling?

\subsection{Experiment Setup}

\noindent\textbf{Datasets.}
We evaluate OMG-VLM on four primary datasets across e-commerce, citation, and social networks. For node classification, we use MAG \textit{Amazon-Arts} \cite{majumder_interview_2020, he_ups_2016} and TAG \textit{ogbn-arXiv} \cite{hu_open_2020}; for link prediction, we use MAG \textit{Amazon-Movies} and IAG \textit{RedditS} \cite{hamilton_inductive_2017}.

To assess cross-graph generalization, we further evaluate all models on held-out graphs unseen during training: TAG \textit{PubMed} \cite{he_harnessing_2024}, TAG \textit{Cora} \cite{mccallum_automating_2000}, MAG \textit{RedditM} \cite{hamilton_inductive_2017}, and MAGs \textit{Amazon-VideoGames} and \textit{Amazon-CDs} \cite{majumder_interview_2020,he_ups_2016}. We mark node classification and link prediction datasets with superscripts $^{\text{\scriptsize NC}}$ and $^{\text{\scriptsize LP}}$, respectively. Dataset statistics are reported in Appendix~\ref{app:dataset_formulation}, and prompt templates are shown in Appendix~\ref{app:prompts_examples}.

\noindent\textbf{Baselines.}
We compare OMG-VLM with five categories of baselines: \textit{MLP} \cite{rosenblatt_perceptron_1958}; GNN-based methods, including \textit{GCN} \cite{kipf_semi-supervised_2017}, \textit{GraphSAGE} \cite{hamilton_inductive_2017}, \textit{RevGAT} \cite{li_training_2021}, and \textit{UniGraph2} \cite{he_unigraph2_2025}; text-only LLM-based methods, including \textit{LLaGA} \cite{chen_llaga_2024} and \textit{GraphPrompter} \cite{lv_graphprompter_2025}; multimodal LLM-based methods, including \textit{MLaGA} \cite{fan_mlaga_2026} and \textit{Graph4MM} \cite{ning_graph4mm_2025}; and VLM-based prompting methods, such as \textit{GraphVLM} \cite{liu2026graphvlm}. We follow standard training and evaluation protocols for all baselines \cite{chen_llaga_2024, lv_graphprompter_2025, fan_mlaga_2026, ning_graph4mm_2025, liu2026graphvlm}.

\noindent\textbf{Evaluation Metric.}
Following prior work \cite{chen_llaga_2024,fan_mlaga_2026,liu2026graphvlm}, we use \textbf{accuracy} as the primary metric for all graph reasoning tasks. We also report additional Macro-F1 results in Appendix~\ref{app:additional_metrics}.

\subsection{Overall Comparison (RQ1 and RQ2)}

We evaluate OMG-VLM under in-domain and transfer settings to assess (i) in-domain effectiveness and (ii) cross-graph generalization across heterogeneous modality schemas.

\noindent\textbf{In-Domain Performance (\textbf{RQ1}).}
\textcircled{1} \textit{\textbf{OMG-VLM achieves the best performance on all in-domain benchmarks across TAG, IAG, and MAG settings.}} As shown in Table~\ref{tab:main_results}, OMG-VLM ranks first on all four training graphs. On TAG arXiv$^{\text{\scriptsize NC}}$, it achieves the highest accuracy, \textbf{67.85 (+3.26 over the best baseline)}, indicating effective use of semantic neighborhood information. The same trend holds on IAG RedditS, showing effectiveness over visual neighborhoods. Gains are especially pronounced on MAGs: OMG-VLM reaches \textbf{92.50} on Arts$^{\text{\scriptsize NC}}$, outperforming the strongest LLM-based baseline LLaGA (88.74) by \textbf{+3.76 points}, and achieves \textbf{95.30} on Movies$^{\text{\scriptsize LP}}$, exceeding UniGraph2 (91.51) by \textbf{+3.79 points}.

\noindent\textbf{Transfer Performance (\textbf{RQ2}).}
\textcircled{2} \textit{\textbf{OMG-VLM generalizes effectively to unseen graphs with strong performance across modality schemas.}} Without dataset-specific fine-tuning, OMG-VLM achieves the best results on most transfer benchmarks. It improves over the strongest baseline by \textbf{+6.42 points} on TAG PubMed$^{\text{\scriptsize NC}}$ and up to \textbf{+20.15 points} on MAG RedditM$^{\text{\scriptsize LP}}$, with consistent gains on all other transfer datasets. The main transfer benchmarks span TAG and MAG settings, while Appendix~\ref{app:flickr_evaluation} additionally evaluates a naturally image-attributed graph, where OMG-VLM outperforms the evaluated baselines. In addition, no baseline consistently ranks second to OMG-VLM in more than three settings. Additional three-seed results for datasets with smaller performance margins are reported in Appendix~\ref{app:multi_seed}.

Overall, OMG-VLM achieves strong in-domain performance and effective transfer to unseen graphs, demonstrating the effectiveness of unified multimodal graph representation learning under heterogeneous modality schemas.

\subsection{Ablation Studies (RQ3)}
In this section, we ablate the proposed aggregation modules and training paradigm.

\noindent\textbf{Effectiveness of Graph Adapters for Neighbor Aggregation.}
We compare the full model with variants that replace (i) textual adapter, (ii) visual adapter, or (iii) both adapters with mean pooling or center-query-conditioned attention pooling, while keeping other components unchanged. Results are detailed in Appendix Table~\ref{tab:adapter_ablation}.

\textit{\textcircled{3} \textbf{Modality-specific graph adapters are crucial for neighborhood aggregation.}}
Replacing both adapters consistently degrades performance. On Movies$^{\text{\scriptsize LP}}$, mean and attention pooling reduce accuracy by \textbf{4.55} and \textbf{21.49} points, respectively; on RedditS$^{\text{\scriptsize LP}}$, both variants incur drops over \textbf{22} points. Replacing only the visual adapter also severely hurts IAGs, reducing RedditS$^{\text{\scriptsize LP}}$ by \textbf{22.85} and \textbf{24.15} points under mean and attention pooling. Replacing the textual adapter lowers performance by up to \textbf{19.30} points on PubMed$^{\text{\scriptsize NC}}$ and \textbf{6.22} points on CDs$^{\text{\scriptsize NC}}$. These results show that simple pooling, even with center-query attention, cannot replace the proposed modality-specific adapters.

\begin{figure}[t]
\centering
\vspace{-0.5em}
\resizebox{0.98\linewidth}{!}{
\begin{tikzpicture}

% ---------- Bars ----------
\begin{axis}[
    width=0.98\linewidth,
    height=4.5cm,
    ybar,
    bar width=7pt,
    ymin=20, ymax=100,
    ylabel={Accuracy (\%)},
    ylabel style={font=\scriptsize},
    symbolic x coords={
        arXiv$^{\text{\scriptsize NC}}$,
        Movies$^{\text{\scriptsize LP}}$,
        RedditS$^{\text{\scriptsize LP}}$,
        RedditM$^{\text{\scriptsize LP}}$,
        CDs$^{\text{\scriptsize NC}}$
    },
    xtick=data,
    xticklabel style={font=\scriptsize, rotate=20, anchor=east},
    yticklabel style={font=\scriptsize},
    ymajorgrids=true,
    grid style={dashed,gray!25},
    axis lines=left,
    enlarge x limits=0.15,
    legend style={
        font=\scriptsize,
        at={(0.5,-0.24)},
        anchor=north,
        legend columns=3,
        draw=none,
        fill=none
    },
]

% Best mean-pooling ablation
\addplot[
    fill=white,
    draw=blue!70!black,
    line width=0.8pt
] coordinates {
(arXiv$^{\text{\scriptsize NC}}$,67.84)
(Movies$^{\text{\scriptsize LP}}$,93.56)
(RedditS$^{\text{\scriptsize LP}}$,97.80)
(RedditM$^{\text{\scriptsize LP}}$,87.15)
(CDs$^{\text{\scriptsize NC}}$,30.89)
};

% Best attention-pooling ablation
\addplot[
    fill=blue!12,
    draw=blue!70!black,
    dashed,
    line width=0.8pt
] coordinates {
(arXiv$^{\text{\scriptsize NC}}$,67.35)
(Movies$^{\text{\scriptsize LP}}$,92.30)
(RedditS$^{\text{\scriptsize LP}}$,97.80)
(RedditM$^{\text{\scriptsize LP}}$,86.85)
(CDs$^{\text{\scriptsize NC}}$,31.53)
};

% Ours
\addplot[
    fill=orange!35,
    draw=orange!85!black,
    line width=1.0pt
] coordinates {
(arXiv$^{\text{\scriptsize NC}}$,67.85)
(Movies$^{\text{\scriptsize LP}}$,95.30)
(RedditS$^{\text{\scriptsize LP}}$,99.10)
(RedditM$^{\text{\scriptsize LP}}$,92.45)
(CDs$^{\text{\scriptsize NC}}$,32.56)
};

\legend{Best Mean, Best Attn., \textbf{Ours}}

\end{axis}

% ---------- Line (Gap) ----------
\begin{axis}[
    width=0.98\linewidth,
    height=4.5cm,
    axis y line*=right,
    axis x line=none,
    ymin=-2, ymax=6,
    ylabel style={font=\scriptsize, xshift=1.8em},
    yticklabel style={font=\scriptsize},
    symbolic x coords={
        arXiv$^{\text{\scriptsize NC}}$,
        Movies$^{\text{\scriptsize LP}}$,
        RedditS$^{\text{\scriptsize LP}}$,
        RedditM$^{\text{\scriptsize LP}}$,
        CDs$^{\text{\scriptsize NC}}$
    },
    xtick=\empty,
]

\addplot[
    color=black,
    mark=o,
    mark size=2.2pt,
    line width=1.0pt,
    mark options={fill=white}
] coordinates {
(arXiv$^{\text{\scriptsize NC}}$,0.01)
(Movies$^{\text{\scriptsize LP}}$,1.74)
(RedditS$^{\text{\scriptsize LP}}$,1.30)
(RedditM$^{\text{\scriptsize LP}}$,5.30)
(CDs$^{\text{\scriptsize NC}}$,1.03)
};

\end{axis}
\end{tikzpicture}
}
\vspace{-0.6em}
\caption{Effectiveness of graph adapters. Bars compare OMG-VLM with the strongest mean-pooling and attention-pooling adapter replacements across textual, visual, or both adapters. The line shows the gap between OMG-VLM and the best pooling variant.}
\label{fig:adapter_ablation_barline}
\vspace{-1.5em}
\end{figure}
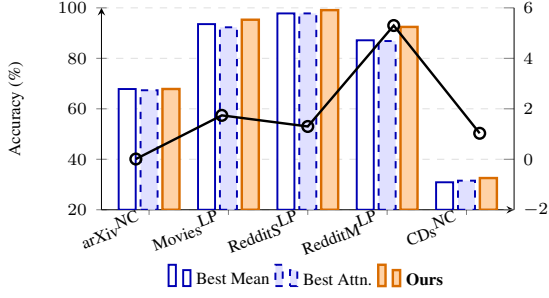
\noindent\textbf{Per-Neighbor Compression in Image Aggregation.}
To evaluate per-neighbor compression, we compare OMG-VLM with and without compression under varying neighborhood sizes, measuring both accuracy and computational cost.

\textit{\textcircled{4} \textbf{Per-neighbor compression improves efficiency while preserving performance.}}
As shown in Appendix Table~\ref{tab:compressor_perf}, compressed and uncompressed variants achieve comparable accuracy across datasets and modality settings, with differences typically within \textbf{1--2 points} (e.g., \textbf{91.17--91.88} vs. \textbf{91.20--92.16} on Arts$^{\text{\scriptsize NC}}$; Figure~\ref{fig:compressor_2x2}). Meanwhile, Figure~\ref{fig:flops_scaling_dense} shows that compression reduces FLOPs by approximately \textbf{75\%} across neighborhood sizes, from \textbf{10.74B--107.37B} to \textbf{2.68B--26.84B}. These results show that per-neighbor compression enables scalable visual neighborhood aggregation with little performance loss.

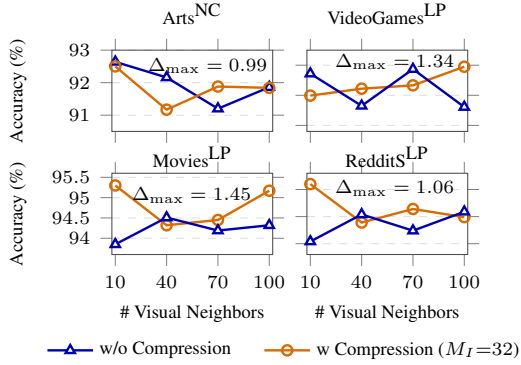
\begin{figure}[t]
\centering
\resizebox{0.94\linewidth}{!}{
\begin{tikzpicture}

\begin{groupplot}[
    group style={
        group size=2 by 2,
        horizontal sep=0.45cm,
        vertical sep=0.55cm,
        xlabels at=edge bottom,
        ylabels at=edge left
    },
    width=0.48\columnwidth,
    height=2.65cm,
    xmin=8, xmax=102,
    xtick={10,40,70,100},
    ymajorgrids=true,
    grid style={dashed,gray!25},
    tick align=outside,
    tick label style={font=\scriptsize},
    label style={font=\scriptsize},
    title style={font=\scriptsize},
    axis line style={black!70},
    clip=false
]

% ------------------ (1) Arts ------------------
\nextgroupplot[
    title={Arts$^{\text{\scriptsize NC}}$},
    ylabel={Accuracy (\%)},
    ymin=90.5, ymax=93.0,
    xticklabels={,,,},
    legend style={
        at={(1.08,-2.45)},
        anchor=north,
        draw=none,
        fill=none,
        font=\scriptsize,
        legend columns=2,
        cells={anchor=west},
        /tikz/every even column/.append style={column sep=10pt}
    },
    legend image post style={line width=0.9pt},
]

\addplot[
    color=blue!70!black,
    mark=triangle*,
    mark size=2.0pt,
    mark options={fill=white},
    line width=0.9pt
] coordinates {(10,92.64)(40,92.16)(70,91.20)(100,91.86)};
\addlegendentry{w/o Compression}

\addplot[
    color=orange!85!black,
    mark=o,
    mark size=1.9pt,
    mark options={fill=white},
    line width=0.9pt
] coordinates {(10,92.50)(40,91.17)(70,91.88)(100,91.84)};
\addlegendentry{w Compression ($M_I{=}32$)}

\node[anchor=north east, font=\scriptsize] at (rel axis cs:1.03,1.03) {$\Delta_{\max}=0.99$};

% ------------------ (2) VG ------------------
\nextgroupplot[
    title={VideoGames$^{\text{\scriptsize LP}}$},
    ymin=76.8, ymax=79.5,
    yticklabels={,,},
    xticklabels={,,,}
]

\addplot[
    color=orange!85!black,
    mark=o,
    mark size=1.9pt,
    mark options={fill=white},
    line width=0.9pt
] coordinates {(10,77.99)(40,78.22)(70,78.33)(100,78.95)};

\addplot[
    color=blue!70!black,
    mark=triangle*,
    mark size=2.0pt,
    mark options={fill=white},
    line width=0.9pt
] coordinates {(10,78.72)(40,77.65)(70,78.87)(100,77.61)};

\node[anchor=north east, font=\scriptsize] at (rel axis cs:0.98,1.08) {$\Delta_{\max}=1.34$};

% ------------------ (3) Movies ------------------
\nextgroupplot[
    title={Movies$^{\text{\scriptsize LP}}$},
    title style={font=\scriptsize, yshift=-0.7em},
    xlabel={\# Visual Neighbors},
    ylabel={Accuracy (\%)},
    ymin=93.6, ymax=95.6
]

\addplot[
    color=orange!85!black,
    mark=o,
    mark size=1.9pt,
    mark options={fill=white},
    line width=0.9pt
] coordinates {(10,95.30)(40,94.32)(70,94.45)(100,95.17)};

\addplot[
    color=blue!70!black,
    mark=triangle*,
    mark size=2.0pt,
    mark options={fill=white},
    line width=0.9pt
] coordinates {(10,93.85)(40,94.51)(70,94.19)(100,94.32)};

\node[anchor=north east, font=\scriptsize] at (rel axis cs:0.93,0.95) {$\Delta_{\max}=1.45$};

% ------------------ (4) RedditS ------------------
\nextgroupplot[
    title={RedditS$^{\text{\scriptsize LP}}$},
    title style={font=\scriptsize, yshift=-0.7em},
    xlabel={\# Visual Neighbors},
    ymin=97.8, ymax=99.3,
    yticklabels={,,}
]

\addplot[
    color=orange!85!black,
    mark=o,
    mark size=1.9pt,
    mark options={fill=white},
    line width=0.9pt
] coordinates {(10,99.10)(40,98.39)(70,98.64)(100,98.49)};

\addplot[
    color=blue!70!black,
    mark=triangle*,
    mark size=2.0pt,
    mark options={fill=white},
    line width=0.9pt
] coordinates {(10,98.04)(40,98.54)(70,98.24)(100,98.59)};

\node[anchor=north east, font=\scriptsize] at (rel axis cs:0.98,1.04) {$\Delta_{\max}=1.06$};

\end{groupplot}

\end{tikzpicture}
}
\vspace{-0.65em}
\caption{Effect of image token compression under increasing visual neighborhood sizes.}
\label{fig:compressor_2x2}
\vspace{-0.8em}
\end{figure}
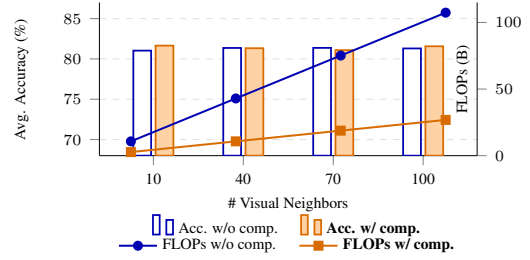
\begin{figure}[t]
\centering

\resizebox{0.92\linewidth}{!}{
\begin{tikzpicture}

% ---------------------- Accuracy (bars) ----------------------
\begin{axis}[
    width=0.98\linewidth,
    height=4.0cm,
    ybar,
    bar width=8pt,
    ymin=68, ymax=87,
    xlabel={\# Visual Neighbors},
    ylabel={Avg. Accuracy (\%)},
    symbolic x coords={10,40,70,100},
    xtick=data,
    ymajorgrids=true,
    grid style={dashed,gray!25},
    tick align=outside,
    tick label style={font=\scriptsize},
    label style={font=\scriptsize},
    axis y line*=left,
    axis x line*=bottom,
    enlarge x limits=0.20,
    legend style={
        font=\scriptsize,
        at={(0.5,-0.35)},
        anchor=north,
        legend columns=2,
        draw=none,
        fill=none,
        /tikz/every even column/.append style={column sep=8pt}
    },
]

\addplot[
    fill=white,
    draw=blue!70!black,
    line width=0.8pt
] coordinates {
(10,81.03)
(40,81.36)
(70,81.37)
(100,81.30)
};
\addlegendentry{Acc. w/o comp.}

\addplot[
    fill=orange!35,
    draw=orange!85!black,
    line width=0.8pt
] coordinates {
(10,81.65)
(40,81.33)
(70,81.09)
(100,81.57)
};
\addlegendentry{\textbf{Acc. w/ comp.}}

\end{axis}

% ---------------------- FLOPs (lines) ----------------------
\begin{axis}[
    width=0.98\linewidth,
    height=4.0cm,
    ymin=0, ymax=115,
    ylabel={FLOPs (B)},
    ylabel style={
        font=\scriptsize,
        at={(axis description cs:1.17,0.5)},
        anchor=center
    },
    yticklabel style={font=\scriptsize},
    yticklabel pos=right,
    axis y line*=right,
    axis x line=none,
    symbolic x coords={10,40,70,100},
    xtick=\empty,
    tick align=outside,
    legend style={
        font=\scriptsize,
        at={(0.5,-0.48)},
        anchor=north,
        legend columns=2,
        draw=none,
        fill=none,
        /tikz/every even column/.append style={column sep=8pt}
    },
]

\addplot[
    color=blue!70!black,
    mark=*,
    mark size=1.8pt,
    line width=0.9pt
] coordinates {
(10,10.74)
(40,42.95)
(70,75.16)
(100,107.37)
};
\addlegendentry{FLOPs w/o comp.}

\addplot[
    color=orange!85!black,
    mark=square*,
    mark size=1.8pt,
    line width=0.9pt
] coordinates {
(10,2.68)
(40,10.74)
(70,18.79)
(100,26.84)
};
\addlegendentry{\textbf{FLOPs w/ comp.}}

\end{axis}

\end{tikzpicture}
}
\vspace{-0.4em}
\caption{Effect of per-neighbor compression as the number of visual neighbors increases. Bars show average accuracy, while lines show computational cost.}
\label{fig:flops_scaling_dense}
\vspace{-0.8em}
\end{figure}
\noindent\textbf{End-to-End Co-Optimization vs. Staged Training.}
We compare end-to-end co-optimization, which jointly optimizes all trainable components, with a two-stage strategy that separates adapter pretraining from structure-aware tuning. Results are shown in Appendix Table~\ref{tab:two_stage_ablation}. We exclude full-parameter fine-tuning due to its substantially higher memory cost, which is infeasible under our hardware setup and less practical than parameter-efficient adaptation \cite{bai_qwen-vl_2023}.

\begin{figure}[t]
\centering
\vspace{-0.3em}
\resizebox{0.94\linewidth}{!}{
\begin{tikzpicture}

% ---------------------- Accuracy (bars) ----------------------
\begin{axis}[
    width=0.98\linewidth,
    height=3.9cm,
    ybar,
    bar width=8pt,
    ymin=70, ymax=100,
    ylabel={Accuracy (\%)},
    symbolic x coords={
        Arts$^{\text{\scriptsize NC}}$,
        Movies$^{\text{\scriptsize LP}}$,
        RedditS$^{\text{\scriptsize LP}}$,
        RedditM$^{\text{\scriptsize LP}}$,
        VideoGames$^{\text{\scriptsize LP}}$
    },
    xtick=data,
    xticklabel style={font=\scriptsize, rotate=20, anchor=east},
    yticklabel style={font=\scriptsize},
    ylabel style={font=\scriptsize},
    ymajorgrids=true,
    grid style={dashed,gray!25},
    axis y line*=left,
    axis x line*=bottom,
    tick align=outside,
    axis line style={black!70},
    enlarge x limits=0.14,
    legend style={
        font=\scriptsize,
        at={(0.5,-0.28)},
        anchor=north,
        legend columns=2,
        draw=none,
        fill=none,
        /tikz/every even column/.append style={column sep=8pt}
    },
    legend cell align=left,
]

\addplot[
    fill=white,
    draw=blue!70!black,
    line width=0.8pt,
] coordinates {
    (Arts$^{\text{\scriptsize NC}}$,92.14)
    (Movies$^{\text{\scriptsize LP}}$,93.97)
    (RedditS$^{\text{\scriptsize LP}}$,96.94)
    (RedditM$^{\text{\scriptsize LP}}$,85.70)
    (VideoGames$^{\text{\scriptsize LP}}$,76.23)
};
\addlegendentry{Two-stage}

\addplot[
    fill=orange!35,
    draw=orange!85!black,
    line width=0.8pt,
] coordinates {
    (Arts$^{\text{\scriptsize NC}}$,92.50)
    (Movies$^{\text{\scriptsize LP}}$,95.30)
    (RedditS$^{\text{\scriptsize LP}}$,99.10)
    (RedditM$^{\text{\scriptsize LP}}$,92.45)
    (VideoGames$^{\text{\scriptsize LP}}$,77.99)
};
\addlegendentry{\textbf{Co-Optimization}}

\end{axis}

% ---------------------- Relative gain (right axis) ----------------------
\begin{axis}[
    width=0.98\linewidth,
    height=3.9cm,
    axis y line*=right,
    axis x line=none,
    ymin=-1, ymax=9,
    ylabel={Gain (\%)},
    ylabel style={
        font=\scriptsize,
        at={(axis description cs:1.17,0.5)},
        anchor=center
    },
    yticklabel style={font=\scriptsize},
    yticklabel pos=right,
    symbolic x coords={
        Arts$^{\text{\scriptsize NC}}$,
        Movies$^{\text{\scriptsize LP}}$,
        RedditS$^{\text{\scriptsize LP}}$,
        RedditM$^{\text{\scriptsize LP}}$,
        VideoGames$^{\text{\scriptsize LP}}$
    },
    xtick=\empty,
    ytick={0,2,4,6,8},
    tick align=outside,
    axis line style={black!70},
    ymajorgrids=false,
]

\addplot[
    color=black,
    mark=o,
    mark size=2.0pt,
    mark options={fill=white},
    line width=0.9pt
] coordinates {
    (Arts$^{\text{\scriptsize NC}}$,0.39)
    (Movies$^{\text{\scriptsize LP}}$,1.41)
    (RedditS$^{\text{\scriptsize LP}}$,2.23)
    (RedditM$^{\text{\scriptsize LP}}$,7.88)
    (VideoGames$^{\text{\scriptsize LP}}$,2.31)
};

\end{axis}

\end{tikzpicture}
}
\vspace{-0.7em}
\caption{Training-strategy ablation. Bars show accuracy under two-stage training and co-optimization; the line shows the relative gain of co-optimization.}
\label{fig:joint_vs_two_stage}
\vspace{-0.8em}
\end{figure}
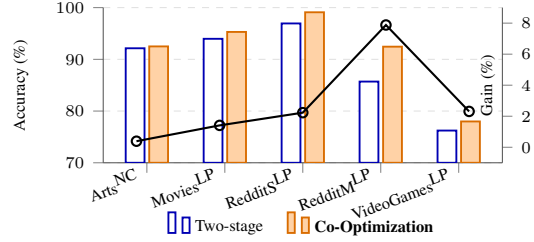

\textit{\textcircled{5} \textbf{End-to-end co-optimization matches or outperforms two-stage training on most datasets while simplifying optimization.}}
Co-optimization performs better on 7 of 9 benchmarks, with gains of \textbf{+1.33} on Movies$^{\text{\scriptsize LP}}$, \textbf{+2.16} on RedditS$^{\text{\scriptsize LP}}$, and \textbf{+6.75} on RedditM$^{\text{\scriptsize LP}}$ (Figure~\ref{fig:joint_vs_two_stage}). It also reduces training time by \textbf{17.54\%} (81.05 $\rightarrow$ 66.83 hours). These results suggest that a unified objective improves co-adaptation between neighborhood modeling and reasoning while avoiding staged training schedules.

\subsection{Sensitivity Analysis (RQ4)}

\noindent\textbf{Injected Text Neighborhood Token Length.}
\textit{\textcircled{6} \textbf{OMG-VLM is stable across different aggregated text token lengths.}}
Table~\ref{tab:hyperparam_sensitivity} of Appendix shows that accuracy varies slightly across $M_T$, typically within \textbf{2 points}. Increasing $M_T$ from 1 to 8 improves performance on several datasets, such as Movies$^{\text{\scriptsize LP}}$ (94.70 $\rightarrow$ \textbf{95.30}) and RedditM$^{\text{\scriptsize LP}}$ (79.10 $\rightarrow$ \textbf{92.45}), while larger values do not consistently help. We set \textbf{$M_T{=}8$} by default, balancing performance with compact textual context.

\noindent\textbf{Per-Neighbor Image Token Length.}
We study sensitivity to the number of compressed visual tokens per neighbor $M_I$, which controls the visual information retained before aggregation.

\textit{\textcircled{7} \textbf{Moderate compression preserves performance, while larger token budgets provide limited gains.}}
As shown in Table~\ref{tab:hyperparam_sensitivity}, accuracy remains stable across $M_I$. For example, Arts$^{\text{\scriptsize NC}}$ stays within \textbf{92.36--92.85}, Movies$^{\text{\scriptsize LP}}$ varies by less than \textbf{0.5} points, suggesting that compact visual summaries preserve key cues for graph reasoning. Aggressive compression ($M_I{=}8$) can remove useful detail, while increasing to $M_I{=}64$ does not consistently improve results. We therefore set \textbf{$M_I{=}32$} by default to keep visual context compact while maintaining strong performance.

\noindent\textbf{Number of Aggregated Neighbors.}
\textit{\textcircled{8} \textbf{OMG-VLM is stable across neighborhood sizes.}}
Table~\ref{tab:hyperparam_sensitivity} shows that accuracy remains relatively stable across $K$. In many cases, $K{=}10$ already achieves the best or near-best performance, e.g., Arts$^{\text{\scriptsize NC}}$ obtains \textbf{92.50} at $K{=}10$ vs.\ 91.17 at $K{=}40$, indicating that a small neighborhood captures most useful structural signals. Larger neighborhoods can occasionally improve transfer performance, e.g., Cora$^{\text{\scriptsize LP}}$ improves from 51.80 to 61.90 at $K{=}70$, but such gains are not consistent. We set \textbf{$K{=}10$} by default for efficient neighborhood aggregation.

\section{Conclusion}
We study attributed graph reasoning under heterogeneous modality schemas, which covers text-, image-, and multimodal-attributed graphs. We propose OMG-VLM, a unified framework that uses a pretrained VLM as a shared backbone and incorporates neighborhood context through structure-aware graph adapters compatible with the backbone embedding space. Experiments on node classification and link prediction across domains and modality schemas show that OMG-VLM consistently outperforms representative baselines and generalizes well to unseen graphs. Ablations further validate the proposed aggregation modules and training strategy. Overall, our results show that VLMs provide a strong foundation for unified attributed graph learning under heterogeneous modality schemas, and we hope this work encourages further exploration of VLM-based graph reasoning.

\section*{Limitations}

OMG-VLM studies unified reasoning over attributed graphs with heterogeneous textual or
visual information. Although this provides a general framework for handling different modality schemas, the current work focuses on static graph reasoning tasks under a standard supervised learning setting. It does not yet consider more interactive paradigms, such as agent-based graph reasoning, where models may dynamically decide how to access, query, or use graph information. Exploring such interactive extensions remains an important direction for future work.

\bibliography{additional, references}

\appendix
\section{Dataset Formulation}
\label{app:dataset_formulation}

\begin{table*}[t]
\scriptsize
\centering
\begingroup
\setlength{\tabcolsep}{2pt}
\renewcommand{\arraystretch}{1.10}
\begin{tabular}{
llrrccc
>{\raggedright\arraybackslash}p{4.7cm}
>{\raggedright\arraybackslash}p{3.3cm}
}
\toprule
\rowcolor{headergray}
\textbf{Setting} &
\textbf{Dataset} &
\textbf{\#Train} &
\textbf{\#Test} &
\textbf{Domain} &
\textbf{Mod.} &
\textbf{Task} &
\textbf{Label space} &
\textbf{Relation / target} \\
\midrule
In-domain
& Amazon-Arts
& 16,917
& 5,639
& Products
& MAG
& NC
& 7 arts-and-crafts categories (e.g., Knitting \& Crochet, Beading \& Jewelry Making)
& Co-purchase / product category \\

& ogbn-arXiv
& 15,545
& 47,416
& Citation
& TAG
& NC
& 40 CS subject areas (e.g., \texttt{cs.AI}, \texttt{cs.CL}, \texttt{cs.CV})
& Paper citation / subject area \\

& Amazon-Movies
& 10,000
& 3,169
& Products
& MAG
& LP
& yes / no
& Co-purchase / edge existence \\

& RedditS
& 6,000
& 1,991
& Social media
& IAG
& LP
& yes / no
& Post interaction / edge existence \\
\midrule
Transfer
& PubMed
& --
& 3,944
& Citation
& TAG
& NC
& 3 diabetes categories (Type 1, Type 2, Experimentally Induced)
& Paper citation / diabetes category \\

& Amazon-CDs
& --
& 7,255
& Products
& MAG
& NC
& 15 music categories (e.g., Jazz, Pop, Blues)
& Co-purchase / product category \\

& RedditM
& --
& 2,000
& Social media
& MAG
& LP
& yes / no
& Post interaction / edge existence \\

& Amazon-VideoGames
& --
& 2,608
& Products
& MAG
& LP
& yes / no
& Co-purchase / edge existence \\

& Cora
& --
& 1,000
& Citation
& TAG
& LP
& yes / no
& Paper citation / edge existence \\
\bottomrule
\end{tabular}
\endgroup
\caption{Dataset details for in-domain and transfer evaluation. Entries in the final column are formatted as graph relation / prediction target. RedditS is treated as an IAG by omitting its textual attributes.}
\label{tab:datasets}
\end{table*}

Table~\ref{tab:datasets} summarizes the datasets used for training and evaluation, including their sizes, domains, modality schemas, tasks, label spaces, graph relations, and prediction targets. The benchmark spans TAGs, IAGs, and MAGs and covers both node classification and link prediction.

All \textbf{transfer datasets} are excluded from training and evaluated without target-dataset fine-tuning. Accordingly, our primary transfer claim concerns cross-graph generalization. Individual datasets additionally test transfer to unseen label spaces, modality-schema shifts, or previously unseen modality-task combinations. For node classification, PubMed and CDs test zero-shot transfer to unseen graphs and label spaces. PubMed additionally introduces a scientific-domain shift, while CDs introduces a product-subdomain shift. For link prediction, the binary task remains unchanged across datasets. VideoGames tests graph and product-subdomain transfer, RedditM introduces an IAG-to-MAG schema shift relative to RedditS, and Cora introduces an unseen TAG-LP combination.

We use the official dataset splits where available, together with the sampling protocol described below. For link prediction, we sample positive edges from the corresponding split and pair them with an equal number of negative edges for training or evaluation. All datasets, pretrained models, and baseline implementations used in this work are publicly available research artifacts, and the associated textual data are in English. We use these artifacts only for research evaluation, follow their standard research-use settings, and adopt protocols consistent with prior work on graph reasoning and multimodal attributed graph learning~\cite{chen_llaga_2024,fan_mlaga_2026,liu2026graphvlm}.

To enable balanced mixed training across tasks, domains, and modality schemas, we downsample selected training graphs to a comparable scale. In particular, we reduce the ogbn-arXiv training split from 101,606 to 15,545 nodes to prevent it from dominating mixed training. Consequently, baseline results reproduced under our unified protocol may differ from originally reported results, which are often obtained under dataset-specific settings. We discuss these differences in Appendix~\ref{app:baseline_details}.

\section{Training Setup and Neighbor Sampling}
\label{app:training_setup}

OMG-VLM is trained for 3 epochs, where the visual encoder is frozen, while the VLM transformer backbone is adapted using LoRA~\cite{hu_lora_2022}. We optimize OMG-VLM with AdamW, using a learning rate of $1\times10^{-5}$, weight decay of 0.1, $\beta_2=0.95$, a warmup ratio of 0.01, and a cosine learning-rate scheduler. For LoRA adaptation, we use a rank of \(r=64\), a scaling factor of \(\alpha=16\), and a dropout rate of 0.05. LoRA is applied to the attention projection modules and feed-forward layers of the VLM backbone.

For \textbf{neighborhood construction} of OMG-VLM, we select neighbors from predefined graph edges for both training and evaluation. Following prior graph-based multimodal learning protocols~\cite{ning_graph4mm_2025}, we randomly sample available 1-hop neighbors up to the neighborhood budget. If capacity remains, we sample 2-hop neighbors and continue to higher hops until the budget is filled. For baselines, we follow the neighborhood construction or sampling strategy specified in the corresponding paper~\cite{fan_mlaga_2026,chen_llaga_2024,ning_graph4mm_2025,he_unigraph2_2025,lv_graphprompter_2025}.

\section{Baseline Details}
\label{app:baseline_details}
Unless otherwise noted, results are reported from single runs conducted under the same experimental setup, using a single NVIDIA H100 80GB GPU. Appendix~\ref{app:multi_seed} reports three-seed evaluations for datasets with smaller performance margins.

\begin{table*}[t]
\centering
\scriptsize
\renewcommand{\arraystretch}{0.01}
\setlength{\abovecaptionskip}{3pt}
\setlength{\belowcaptionskip}{-4pt}

\resizebox{\textwidth}{!}{
\begin{tabular}{lcccccc}
\toprule
\rowcolor{headergray}
\textbf{Dataset}
& \textbf{LLaGA}
& \textbf{GraphPrompter}
& \textbf{UniGraph2}
& \textbf{Graph4MM}
& \textbf{GraphVLM}
& \textbf{OMG-VLM} \\
\midrule
Arts$^{\text{\scriptsize NC}}$
& \underline{88.51} & 17.04 & 79.60 & 85.64 & 66.19 & \textbf{92.13} \\

ogbn-arXiv$^{\text{\scriptsize NC}}$
& \underline{74.71} & 70.83 & 32.84 & 67.14 & 42.52 & \textbf{76.53} \\

Movies$^{\text{\scriptsize LP}}$
& 52.13 & 18.18 & \underline{91.51} & 75.39 & 49.10 & \textbf{95.08} \\

RedditS$^{\text{\scriptsize LP}}$
& -- & -- & 65.24 & -- & \underline{90.81} & \textbf{99.40} \\

PubMed$^{\text{\scriptsize NC}}$
& 1.52 & 11.59 & 21.70 & 1.01 & \underline{63.09} & \textbf{75.99} \\

RedditM$^{\text{\scriptsize LP}}$
& 44.50 & 8.95 & 61.25 & \underline{73.90} & 47.00 & \textbf{87.50} \\

VideoGames$^{\text{\scriptsize LP}}$
& 46.36 & 0.50 & 49.08 & \underline{63.61} & 49.15 & \textbf{77.84} \\

CDs$^{\text{\scriptsize NC}}$
& 4.88 & 11.83 & 5.97 & 0.44 & \underline{33.65} & \textbf{34.29} \\

Cora$^{\text{\scriptsize LP}}$
& 1.70 & 1.40 & 51.80 & 52.60 & \underline{62.60} & \textbf{69.10} \\
\bottomrule
\end{tabular}
}

\caption{Additional reproduced baseline results, with ogbn-arXiv$^{\text{\scriptsize NC}}$ no longer downsampled. -- indicates that the model cannot process the dataset due to architectural constraints or unavailable results. Bold and underline indicate the best and second-best results.}
\label{tab:full_arxiv_baseline}
\end{table*}

\subsection{Unavailable Baseline Results}
\label{app:unavailable_baselines}

Certain baseline results are unavailable because the corresponding models are designed for specific modality schemas and cannot be directly applied to all dataset types considered in our evaluation.

\textbf{Text-only LLM-based baselines}, including \textbf{LLaGA}~\cite{chen_llaga_2024} and \textbf{GraphPrompter}~\cite{lv_graphprompter_2025}, are designed for TAGs. Their input construction relies on textual node attributes and serialized graph context, without a mechanism for processing visual node attributes. Therefore, they cannot be applied to RedditS, which is an IAG containing only visual inputs.

\textbf{Multimodal LLM-based baselines} also have architecture-specific modality requirements. \textbf{Graph4MM}~\cite{ning_graph4mm_2025} is designed for MAGs and uses an MM-QFormer module to extract visual information through text-derived queries. This design requires textual attributes and, therefore, cannot be directly applied to IAGs without text. Conversely, \textbf{MLaGA}~\cite{fan_mlaga_2026} reasons over graph structure using a visual branch and a multimodal fusion module, which assumes the availability of visual inputs. As a result, it is not directly applicable to TAGs, where only textual attributes are available.

These unavailable results highlight the motivation for our heterogeneous modality setting. Unlike prior methods that are specialized for TAGs, IAGs, or MAGs, OMG-VLM is designed as a unified framework that can operate across graphs with different modality schemas.

\subsection{Differences from Reported Baselines}
\label{app:baseline_reproduction}

Some reproduced baseline results differ from those reported in the original papers, particularly MLaGA on Movies$^{\text{\scriptsize LP}}$ and LLaGA on ogbn-arXiv$^{\text{\scriptsize NC}}$. These gaps mainly stem from differences between prior dataset-specific protocols and our unified heterogeneous-modality benchmark. Our setting trains and evaluates models across heterogeneous graphs, including TAGs, IAGs, and MAGs, whereas many baselines are designed for narrower modality settings, such as LLaGA for TAGs and MLaGA for MAGs~\cite{chen_llaga_2024,fan_mlaga_2026}. Our protocol also jointly mixes citation, e-commerce, and social graphs, rather than training within a single domain or modality schema.

\textbf{MLaGA.}
For MLaGA, the gap on Movies$^{\text{\scriptsize LP}}$ is likely related to link-prediction split construction. Since MLaGA does not release the exact train/test edge splits, our reproduction may use different sampled edges for training and evaluation. Moreover, MLaGA is originally evaluated in a narrower Amazon-domain MAG setting~\cite{fan_mlaga_2026}, while our protocol covers heterogeneous domains and modality schemas.

\textbf{LLaGA.}
For LLaGA, the gap on arXiv$^{\text{\scriptsize NC}}$ mainly comes from the balanced mixed-training setup. To prevent large graphs from dominating joint training, we downsample ogbn-arXiv training nodes, as detailed in Appendix~\ref{app:dataset_formulation}, which differs from the original LLaGA setting, which uses the full ogbn-arXiv split~\cite{chen_llaga_2024}. To isolate the effect of training-set size, we additionally evaluate reproduced baselines with the full ogbn-arXiv training data. As shown in Table~\ref{tab:full_arxiv_baseline}, LLaGA improves to 74.71 on ogbn-arXiv, while OMG-VLM further improves to 76.53, suggesting that the lower reproduced LLaGA result in Table~\ref{tab:main_results} is \textbf{largely attributable to the balanced split rather than implementation changes}. Meanwhile, OMG-VLM remains the best-performing model across datasets, outperforming LLaGA by 1.82 points.

Overall, these results indicate that differences from originally reported baselines mainly arise from adapting specialized methods to a unified benchmark over heterogeneous domains and modality schemas. Meanwhile, OMG-VLM shows competitive performance in both training settings.

\subsection{Additional Evaluation Metrics}
\label{app:additional_metrics}
\begin{table}[t]
\centering
\scriptsize
\setlength{\tabcolsep}{3.2pt}
\renewcommand{\arraystretch}{0.9}
\setlength{\abovecaptionskip}{3pt}
\setlength{\belowcaptionskip}{-4pt}

\resizebox{\columnwidth}{!}{
\begin{tabular}{lccccc}
\toprule
\rowcolor{headergray}
\textbf{Dataset}
& \textbf{GraphP.}
& \textbf{UniG2}
& \textbf{G4MM}
& \textbf{G-VLM}
& \textbf{OMG} \\
\midrule
Arts$^{\text{\scriptsize NC}}$
& 84.81 & 80.78 & \underline{87.87} & 80.06 & \textbf{93.21} \\

arXiv$^{\text{\scriptsize NC}}$
& 23.68 & 22.82 & 37.39 & \underline{52.94} & \textbf{54.07} \\

Movies$^{\text{\scriptsize LP}}$
& 15.34 & \underline{91.46} & 85.95 & 33.15 & \textbf{95.19} \\

RedditS$^{\text{\scriptsize LP}}$
& -- & \underline{78.85} & -- & 54.90 & \textbf{99.10} \\

PubMed$^{\text{\scriptsize NC}}$
& 43.80 & 38.40 & 33.13 & \underline{59.64} & \textbf{73.34} \\

RedditM$^{\text{\scriptsize LP}}$
& 37.00 & 62.35 & \underline{75.32} & 46.53 & \textbf{92.14} \\

CDs$^{\text{\scriptsize NC}}$
& 22.89 & 5.15 & 12.05 & \underline{26.34} & \textbf{27.07} \\
\bottomrule
\end{tabular}
}

\caption{Macro-F1 results for representative strong baselines and OMG-VLM. -- indicates that the model cannot process the dataset due to architectural constraints. Bold and underline indicate the best and second-best results.}
\label{tab:macro_f1}
\end{table}
We report \textbf{Macro-F1} for representative strong baselines and OMG-VLM in Table~\ref{tab:macro_f1}. Macro-F1 provides a complementary view of performance by weighting classes equally, which is especially useful when class distributions are imbalanced. OMG-VLM achieves the best Macro-F1 on most datasets, with particularly large gains on PubMed$^{\text{\scriptsize NC}}$ (\textbf{73.34} vs.\ 59.64), and RedditM$^{\text{\scriptsize LP}}$ (\textbf{92.14} vs.\ 75.32). These results are consistent with the accuracy trends in the main paper and further support the effectiveness of OMG-VLM across tasks and modality schemas.

\subsection{Evaluation on a Naturally Image-Attributed Graph}
\label{app:flickr_evaluation}

RedditS provides a controlled image-only setting constructed by omitting its textual attributes. To complement this setting, we evaluate on the naturally image-attributed Flickr graph distributed with GraphSAINT~\cite{zeng_graphsaint_2020}. The graph originates from NUS-WIDE~\cite{chua_nus-wide_2009} and contains 89,250 image nodes, 899,756 edges, and seven single-label classes. Each node represents a Flickr image, while edges connect images sharing properties such as geographic location, gallery membership, or user interactions. We match the evaluated nodes to their corresponding raw NUS-WIDE images. Flickr is excluded from training, and all models are evaluated on its 22,313-node test split without target-dataset fine-tuning. We compare OMG-VLM with representative baselines from the main evaluation.

\begin{table}[t]
\small
\centering
\begin{tabular}{lc}
\toprule
\rowcolor{headergray}
\textbf{Model} & \textbf{Accuracy (\%)} \\
\midrule
MLP & 7.38 \\
GCN & 25.47 \\
UniGraph2 & 35.82 \\
GraphVLM & 25.79 \\
\textbf{OMG-VLM} & \textbf{42.29} \\
\bottomrule
\end{tabular}
\caption{Transfer performance of OMG-VLM and representative baselines from the main evaluation on the naturally image-attributed Flickr graph. All models are evaluated on the 22,313-node test split without Flickr-specific fine-tuning. Bold indicates the best result.}
\vspace{-1em}
\label{tab:flickr_native_iag}
\end{table}

As shown in Table~\ref{tab:flickr_native_iag}, OMG-VLM achieves an accuracy of \textbf{42.29}, outperforming UniGraph2 by \textbf{6.47 points}. This result supports OMG-VLM's ability to leverage visual graph neighborhoods and transfer to unseen IAGs.

\subsection{Multi-Seed Evaluation}
\label{app:multi_seed}

To assess the robustness of comparisons with smaller single-run margins, we rerun OMG-VLM and the strongest corresponding baseline on arXiv, CDs, and Cora using three random seeds under the same experimental setup. Table~\ref{tab:multi_seed} reports the mean accuracy and standard deviation.

\begin{table}[t]
\small
\centering
\begin{tabular}{lcc}
\toprule
\rowcolor{headergray}
\textbf{Dataset} & \textbf{Strongest Baseline} & \textbf{OMG-VLM} \\
\midrule
arXiv$^{\text{\scriptsize NC}}$
& $64.92 \pm 0.46$ (MLP)
& $\mathbf{67.95 \pm 0.04}$ \\
CDs$^{\text{\scriptsize NC}}$
& $30.86 \pm 0.43$ (MLaGA)
& $\mathbf{32.29 \pm 0.24}$ \\
Cora$^{\text{\scriptsize LP}}$
& $50.63 \pm 0.63$ (UniGraph2)
& $\mathbf{51.67 \pm 0.15}$ \\
\bottomrule
\end{tabular}
\caption{Three-seed accuracy on datasets with smaller single-run performance margins. Values are reported as mean $\pm$ standard deviation, with the strongest baseline identified in parentheses. Bold indicates the best results.}
\label{tab:multi_seed}
\end{table}

As shown in Table~\ref{tab:multi_seed}, OMG-VLM achieves \textbf{higher mean accuracy} on all three datasets. Its improvements of \textbf{3.03 points} on arXiv, \textbf{1.43 points} on CDs, and \textbf{1.04 points} on Cora are larger than the observed run-to-run variation, supporting the consistency of the single-run results.

\section{Computational Cost Analysis}
\label{app:computational_cost}
\begin{table}[t]
\centering
\scriptsize
\setlength{\tabcolsep}{3.2pt}
\renewcommand{\arraystretch}{1.2}
\setlength{\abovecaptionskip}{3pt}
\setlength{\belowcaptionskip}{-4pt}

\resizebox{\columnwidth}{!}{
\begin{tabular}{lccccc}
\toprule
\rowcolor{headergray}
\textbf{Model}
& \textbf{Backbone}
& \textbf{Train}
& \textbf{Infer}
& \textbf{Avg. Prompt} \\
\rowcolor{headergray}
\textbf{}
& \textbf{}
& \textbf{(ms/sp)}
& \textbf{(ms/sp)}
& \textbf{Tokens} \\
\midrule

GraphVLM
& VLM
& 1428.30
& 397.85
& 1444.07 \\

OMG-VLM
& VLM
& \textbf{925.42}
& \textbf{358.17}
& \textbf{464.30} \\
\bottomrule
\end{tabular}
}

\caption{Computation comparison. Runtime is reported in milliseconds per sample. \textit{Avg. Prompt} denotes the average number of input prompt tokens of Arts$^{\text{\scriptsize NC}}$. Bold highlights the better result among VLM-based methods.}
\label{tab:computational_cost}
\end{table}
We report the computational cost of representative multimodal baselines in Table~\ref{tab:computational_cost}. All results are measured on a single NVIDIA H100 80GB GPU under comparable configurations. Since OMG-VLM is built on a VLM backbone, the most meaningful efficiency comparison is with VLM-based graph reasoning methods rather than LLM-only or GNN-based baselines, whose computational profiles are fundamentally different. We therefore focus our discussion on GraphVLM, which represents an intuitive VLM-based solution that incorporates neighborhood information by directly flattening graph context into the prompt.

\begin{table*}[t]
\centering
\scriptsize
\setlength{\abovecaptionskip}{3pt}
\setlength{\belowcaptionskip}{-4pt}

\begin{tabular}{lcc|cc|cc|c}
\toprule
\rowcolor{headergray}
\textbf{Dataset}
& \multicolumn{2}{c|}{\textbf{Text Pool}}
& \multicolumn{2}{c|}{\textbf{Image Pool}}
& \multicolumn{2}{c|}{\textbf{Text/Image Pool}}
& \textbf{Full} \\
\rowcolor{headergray}
& \textbf{Mean} & \textbf{Attn.}
& \textbf{Mean} & \textbf{Attn.}
& \textbf{Mean} & \textbf{Attn.}
& \textbf{Model} \\
\midrule
Arts$^{\text{\scriptsize NC}}$
& 92.43 (-0.07) & 92.20 (-0.30)
& 92.30 (-0.20) & 91.25 (-1.25)
& 92.20 (-0.30) & 91.67 (-0.83)
& \textbf{92.50} \\

arXiv$^{\text{\scriptsize NC}}$
& 67.38 (-0.47) & 67.09 (-0.76)
& 67.84 (-0.01) & 67.35 (-0.50)
& 67.12 (-0.73) & 67.09 (-0.76)
& \textbf{67.85} \\

Movies$^{\text{\scriptsize LP}}$
& 92.93 (-2.37) & 92.30 (-3.00)
& 93.56 (-1.74) & 85.33 (-9.97)
& 90.75 (-4.55) & 73.81 (-21.49)
& \textbf{95.30} \\

RedditS$^{\text{\scriptsize LP}}$
& 97.80 (-1.30) & 97.80 (-1.30)
& 76.25 (-22.85) & 74.95 (-24.15)
& 76.30 (-22.80) & 74.70 (-24.40)
& \textbf{99.10} \\

PubMed$^{\text{\scriptsize NC}}$
& 59.25 (-19.30) & 60.41 (-18.14)
& 78.27 (-0.28) & 78.32 (-0.23)
& 57.13 (-21.42) & 56.25 (-23.30)
& \textbf{78.55} \\

RedditM$^{\text{\scriptsize LP}}$
& 87.15 (-5.30) & 86.85 (-5.60)
& 79.85 (-12.60) & 82.30 (-10.15)
& 80.70 (-11.75) & 76.45 (-16.00)
& \textbf{92.45} \\

VideoGames$^{\text{\scriptsize LP}}$
& 76.57 (-1.42) & 76.99 (-1.00)
& \textbf{78.26 (+0.27)} & 65.57 (-12.42)
& 69.75 (-8.24) & 65.49 (-12.50)
& 77.99 \\

CDs$^{\text{\scriptsize NC}}$
& 26.34 (-6.22) & 31.53 (-1.03)
& 30.89 (-1.67) & 27.21 (-5.35)
& 26.34 (-6.22) & 30.76 (-1.80)
& \textbf{32.56} \\

Cora$^{\text{\scriptsize LP}}$
& 51.00 (-0.80) & 50.50 (-1.30)
& 51.49 (-0.31) & \textbf{54.50 (+2.70)}
& 50.90 (-0.90) & 50.60 (-1.20)
& 51.80 \\
\bottomrule
\end{tabular}

\caption{Ablation on neighborhood aggregation modules. Mean replaces the corresponding adapter with mean pooling over neighbor representations, while Attn. uses center-query attention pooling, where the center node representation attends to neighbors. Text/Image Pool replaces both adapters. Parenthesized values are signed accuracy-point differences from the full OMG-VLM model.}
\label{tab:adapter_ablation}
\end{table*}

Compared with GraphVLM, OMG-VLM achieves lower training time, inference time, and prompt length. The main reason is that GraphVLM serializes graph neighborhoods directly into the textual prompt, so its input length grows with the number of neighbors and induces a higher attention cost. In contrast, OMG-VLM compresses graph-derived context into a fixed number of tokens before passing it to the VLM. This bounded context reduces prompt expansion while retaining VLM-based multimodal reasoning capacity.

\section{Ablation and Sensitivity Results}
\label{app:ablation}

\subsection{Complete Main Ablation Results}
\label{app:main_ablation_results}

\begin{table}[t]
\centering
\scriptsize
\setlength{\tabcolsep}{2.8pt}
\renewcommand{\arraystretch}{0.88}
\setlength{\abovecaptionskip}{3pt}
\setlength{\belowcaptionskip}{-4pt}

\resizebox{\columnwidth}{!}{
\begin{tabular}{lcccc|cccc}
\toprule
\rowcolor{headergray}
\textbf{Dataset}
& \multicolumn{4}{c|}{\textbf{w/ Compr.}}
& \multicolumn{4}{c}{\textbf{w/o Compr.}} \\
\rowcolor{headergray}
& \textbf{10} & \textbf{40} & \textbf{70} & \textbf{100}
& \textbf{10} & \textbf{40} & \textbf{70} & \textbf{100} \\
\midrule
Arts$^{\text{\scriptsize NC}}$
& 92.50 & 91.17 & 91.88 & 91.84
& 92.64 & 92.16 & 91.20 & 91.86 \\

Movies$^{\text{\scriptsize LP}}$
& 95.30 & 94.32 & 94.45 & 95.17
& 93.85 & 94.51 & 94.19 & 94.32 \\

RedditS$^{\text{\scriptsize LP}}$
& 99.10 & 98.39 & 98.64 & 98.49
& 98.04 & 98.54 & 98.24 & 98.59 \\

RedditM$^{\text{\scriptsize LP}}$
& 92.45 & 92.40 & 89.60 & 93.10
& 89.35 & 91.30 & 91.05 & 90.85 \\

VideoGames$^{\text{\scriptsize LP}}$
& 77.99 & 78.22 & 78.37 & 78.95
& 78.72 & 77.65 & 78.87 & 77.61 \\

CDs$^{\text{\scriptsize NC}}$
& 32.56 & 33.46 & 33.58 & 31.85
& 33.58 & 33.98 & 34.69 & 34.57 \\

\bottomrule
\end{tabular}
}

\caption{Ablation on per-neighbor image compression across visual neighborhood sizes. Compressed visual token length is fixed to $M_I{=}32$.}
\label{tab:compressor_perf}
\end{table}

\begin{table}[t]
\centering
\scriptsize
\setlength{\tabcolsep}{3.5pt}
\renewcommand{\arraystretch}{1}
\setlength{\abovecaptionskip}{3pt}
\setlength{\belowcaptionskip}{-4pt}

\begin{tabular}{@{}lcc@{}}
\toprule
\rowcolor{headergray}
\textbf{Dataset} & \textbf{Co-Optimization} & \textbf{Two-Stage} \\
\midrule
Arts$^{\text{\scriptsize NC}}$       & \textbf{92.50} & 92.14 \\
arXiv$^{\text{\scriptsize NC}}$      & \textbf{67.85} & 66.65 \\
Movies$^{\text{\scriptsize LP}}$     & \textbf{95.30} & 93.97 \\
RedditS$^{\text{\scriptsize LP}}$    & \textbf{99.10} & 96.94 \\
PubMed$^{\text{\scriptsize NC}}$     & 78.55          & \textbf{90.42} \\
RedditM$^{\text{\scriptsize LP}}$    & \textbf{92.45} & 85.70 \\
VideoGames$^{\text{\scriptsize LP}}$ & \textbf{77.99} & 76.23 \\
CDs$^{\text{\scriptsize NC}}$        & \textbf{32.56} & 31.63 \\
Cora$^{\text{\scriptsize LP}}$       & 51.80          & \textbf{66.20} \\
\bottomrule
\end{tabular}

\caption{Performance of OMG-VLM under end-to-end co-optimization and two-stage training. Bold indicates the best results.}
\vspace{-1em}
\label{tab:two_stage_ablation}
\end{table}

Table~\ref{tab:adapter_ablation} details the performance of OMG-VLM, text-pooled variant, image-pooled variant, and both-pooled variant of our model on all datasets.

\noindent Table~\ref{tab:compressor_perf} details the performance of OMG-VLM with and without per-neighbor compression across different neighborhood sizes (10-100).

\noindent Table~\ref{tab:two_stage_ablation} details the performance of OMG-VLM under co-optimization and two-stage training.

\subsection{Complete Sensitivity Results}

\noindent Table~\ref{tab:hyperparam_sensitivity} details the performance of OMG-VLM with different textual token lengths, visual token lengths, and the number of aggregated neighbors.

\begin{table*}[t]
\centering
\scriptsize
\renewcommand{\arraystretch}{0.001}
\setlength{\abovecaptionskip}{3pt}
\setlength{\belowcaptionskip}{-4pt}

\resizebox{\textwidth}{!}{
\begin{tabular}{llcccc|ccc|cccc}
\toprule
\rowcolor{headergray}
\textbf{Setting} & \textbf{Dataset}
& \multicolumn{4}{c}{\textbf{Text Tokens} ($M_T$)}
& \multicolumn{3}{c}{\textbf{Image Tokens} ($M_I$)}
& \multicolumn{4}{c}{\textbf{Neighbors} ($K$)} \\
\rowcolor{headergray}
& & \multicolumn{4}{c}{$M_I=32, K=10$}
& \multicolumn{3}{c}{$M_T=1, K=10$}
& \multicolumn{4}{c}{$M_T=8, M_I=32$} \\
\cmidrule(lr){3-6} \cmidrule(lr){7-9} \cmidrule(lr){10-13}
& & 1 & 8 & 16 & 32
& 8 & 32 & 64
& 10 & 40 & 70 & 100 \\
\midrule

\multirow{4}{*}{\textbf{In-Domain}}
& Arts$^{\text{\scriptsize NC}}$
& \textbf{92.73} & 92.50 & 92.36 & 92.04
& 92.36 & \textbf{92.73} & 92.85
& \textbf{92.50} & 91.17 & 91.88 & 91.84 \\
& arXiv$^{\text{\scriptsize NC}}$
& 68.14 & 67.85 & 67.98 & \textbf{68.16}
& 68.06 & \textbf{68.14} & 67.67
& 67.85 & 67.91 & 67.89 & \textbf{67.93} \\
& Movies$^{\text{\scriptsize LP}}$
& 94.70 & \textbf{95.30} & 94.45 & 94.76
& \textbf{94.98} & 94.70 & 94.67
& \textbf{95.30} & 94.32 & 94.45 & 95.17 \\
& RedditS$^{\text{\scriptsize LP}}$
& 98.95 & \textbf{99.10} & 98.64 & 98.19
& 98.69 & \textbf{98.95} & 98.90
& \textbf{99.10} & 98.39 & 98.64 & 98.49 \\
\midrule

\multirow{5}{*}{\textbf{Transfer}}
& PubMed$^{\text{\scriptsize NC}}$
& 77.28 & 78.55 & \textbf{77.87} & 77.05
& \textbf{78.60} & 77.28 & 77.08
& 78.55 & 76.17 & \textbf{78.72} & 76.98 \\
& RedditM$^{\text{\scriptsize LP}}$
& 79.10 & \textbf{92.45} & 83.80 & 89.85
& \textbf{91.55} & 79.10 & 86.95
& 92.45 & 92.40 & 89.60 & \textbf{93.10} \\
& VideoGames$^{\text{\scriptsize LP}}$
& 78.14 & 77.99 & 74.08 & \textbf{78.49}
& 77.72 & 78.14 & \textbf{79.41}
& 77.99 & 78.22 & 78.37 & \textbf{78.95} \\
& CDs$^{\text{\scriptsize NC}}$
& 32.49 & 32.56 & 32.39 & \textbf{35.22}
& 32.10 & 32.49 & \textbf{34.60}
& 32.56 & 33.46 & \textbf{33.58} & 31.85 \\
& Cora$^{\text{\scriptsize LP}}$
& \textbf{56.30} & 51.80 & 50.50 & 51.00
& 52.00 & \textbf{56.30} & 52.70
& 51.80 & 52.30 & \textbf{61.90} & 55.70 \\

\bottomrule
\end{tabular}
}
\caption{Sensitivity analysis to textual token length ($M_T$), visual token length ($M_I$), and number of aggregated neighbors ($K$). Bold indicates the best results.}
\label{tab:hyperparam_sensitivity}
\end{table*}
\begin{table*}[t]
\centering
\scriptsize
\setlength{\tabcolsep}{3.4pt}
\renewcommand{\arraystretch}{0.92}
\setlength{\abovecaptionskip}{3pt}
\setlength{\belowcaptionskip}{-4pt}

\resizebox{\textwidth}{!}{
\begin{tabular}{lccccccccc}
\toprule
\rowcolor{headergray}
\textbf{Backbone Model}
& \textbf{Arts$^{\text{\scriptsize NC}}$}
& \textbf{arXiv$^{\text{\scriptsize NC}}$}
& \textbf{Movies$^{\text{\scriptsize LP}}$}
& \textbf{RedditS$^{\text{\scriptsize LP}}$}
& \textbf{PubMed$^{\text{\scriptsize NC}}$}
& \textbf{CDs$^{\text{\scriptsize NC}}$}
& \textbf{RedditM$^{\text{\scriptsize LP}}$}
& \textbf{VideoGames$^{\text{\scriptsize LP}}$} \\
\midrule

Best Baseline
& 89.63
& 64.59
& 91.51
& \underline{90.10}
& 72.13
& 31.10
& 72.30
& 70.30\\

OMG-VLM(LLaVA-v1.6-7B)
& \underline{91.36}
& 66.30
& 93.12
& 89.05
& \underline{88.97}
& \underline{34.73}
& \underline{81.50}
& \underline{76.93}\\

OMG-VLM(Qwen3-VL-4B)
& 89.00
& \underline{67.67}
& \underline{93.66}
& 83.31
& \textbf{91.78}
& \textbf{40.88}
& 76.65
& 76.29\\

\textbf{OMG-VLM(Qwen-VL-7B)}
& \textbf{92.50}
& \textbf{67.85}
& \textbf{95.30}
& \textbf{99.10}
& 78.55
& 32.56
& \textbf{92.45}
& \textbf{77.99}
\\

\bottomrule
\end{tabular}
}

\caption{Additional backbone ablation results. We instantiate OMG-VLM with different VLM backbones and compare them with the strongest non-OMG-VLM baseline on each dataset. Bold indicates the best result, and underline indicates the second-best result.}
\label{tab:backbone_ablation}
\end{table*}

\subsection{Backbone Ablations}
\label{app:backbone_ablation}

The main experiments use the exact Hugging Face checkpoint \path{Qwen/Qwen-VL-Chat}, which contains approximately 7B parameters~\cite{bai_qwen-vl_2023}. We select this backbone because it matches the model version and scale used by GraphVLM~\cite{liu2026graphvlm}, the closest VLM-based baseline. This controlled choice allows the comparison to focus on graph adaptation rather than improvements from using a larger or more recent backbone.

To evaluate sensitivity to the backbone, we instantiate the same OMG-VLM framework with \path{llava-hf/llava-v1.6-vicuna-7b-hf}~\cite{liu2024llavanext}, which represents a different VLM family at a comparable 7B scale, and \path{Qwen/Qwen3-VL-4B-Instruct}~\cite{bai_qwen3-vl_2025}, which represents a smaller model from the Qwen VLM family. The graph-adapter design and evaluation protocol remain unchanged across these variants. Table~\ref{tab:backbone_ablation} reports their results alongside the strongest non-OMG-VLM baseline on each dataset.

Overall, OMG-VLM remains effective across backbone families and model scales. The Qwen-VL-based version achieves the best results on most datasets, including \textbf{95.30} on Movies$^{\text{\scriptsize LP}}$, \textbf{99.10} on RedditS$^{\text{\scriptsize LP}}$, and \textbf{92.45} on RedditM$^{\text{\scriptsize LP}}$, substantially outperforming the strongest corresponding baselines of 91.51, 90.10, and 72.30. The alternative backbones also remain competitive: LLaVA-v1.6-7B-based variant reaches 91.36 on Arts$^{\text{\scriptsize NC}}$ and 93.12 on Movies$^{\text{\scriptsize LP}}$, while Qwen3-VL-4B-based variant achieves 91.78 on PubMed$^{\text{\scriptsize NC}}$ and 40.88 on CDs$^{\text{\scriptsize NC}}$. Although Qwen3-VL-4B trails the main Qwen-VL-7B variant on several datasets, it differs in both architecture and parameter scale and still exceeds the strongest non-OMG-VLM baseline on six of eight datasets. These results suggest that OMG-VLM is not tied to a particular pretrained VLM implementation. Instead, its graph adapters can operate on the native textual and visual representations of different VLM backbones without backbone-specific redesign. Together, these results indicate that OMG-VLM's effectiveness is not restricted to a single VLM family or parameter scale.

\subsection{Fine-Tuned and Zero-Shot VLM Controls}
\label{app:vlm_controls}

\noindent\textbf{Matched Fine-Tuned No-Graph Control.}
To distinguish the contribution of graph neighborhoods from target-attribute understanding and task adaptation, we construct a matched no-graph control using the same \texttt{Qwen/Qwen-VL-Chat} backbone as OMG-VLM. We fine-tune this control on the same training examples and task instructions while removing all textual and visual neighborhood inputs.

\noindent\textbf{Zero-Shot Controls.}
We further examine whether the gains can be obtained by prompting the pretrained \texttt{Qwen/Qwen-VL-Chat} checkpoint without graph-adapter training. In the non-structural setting, the model receives only the available target-node attributes and the task instruction. In the structural setting, it additionally receives neighborhood information serialized into the prompt. Neither zero-shot setting updates any parameters.

\begin{table}[t]
\centering
\scriptsize
\setlength{\tabcolsep}{2.4pt}
\renewcommand{\arraystretch}{0.92}
\setlength{\abovecaptionskip}{3pt}
\setlength{\belowcaptionskip}{-4pt}

\begin{tabular}{lcccc}
\toprule
\rowcolor{headergray}
\textbf{Dataset}
& \multicolumn{2}{c}{\textbf{Zero-Shot}}
& \shortstack{\textbf{FT}\\\textbf{No-Graph}}
& \textbf{OMG-VLM} \\
\rowcolor{headergray}
& \textbf{Non-Struct.}
& \textbf{Struct.}
& & \\
\midrule
Arts$^{\text{\scriptsize NC}}$
& 69.39 & 62.09 & \underline{74.61} & \textbf{92.50} \\

arXiv$^{\text{\scriptsize NC}}$
& 2.89 & 8.50 & \underline{44.62} & \textbf{67.85} \\

Movies$^{\text{\scriptsize LP}}$
& 61.15 & 50.05 & \underline{83.02} & \textbf{95.30} \\

RedditS$^{\text{\scriptsize LP}}$
& 49.90 & 62.68 & \underline{85.38} & \textbf{99.10} \\

PubMed$^{\text{\scriptsize NC}}$
& \underline{77.32} & 72.61 & 72.99 & \textbf{78.55} \\

RedditM$^{\text{\scriptsize LP}}$
& \underline{68.60} & 38.95 & 67.75 & \textbf{92.45} \\

VideoGames$^{\text{\scriptsize LP}}$
& 50.08 & 50.00 & \underline{64.92} & \textbf{77.99} \\

CDs$^{\text{\scriptsize NC}}$
& 28.73 & 29.20 & \underline{30.70} & \textbf{32.56} \\

Cora$^{\text{\scriptsize LP}}$
& 50.00 & \underline{50.50} & \underline{50.50} & \textbf{51.80} \\
\midrule
\textbf{Average}
& 50.90 & 47.18 & \underline{63.83} & \textbf{76.46} \\
\bottomrule
\end{tabular}

\caption{Accuracy of OMG-VLM and Qwen-VL controls. Non-Struct.\ and Struct.\ denote zero-shot prompting without and with serialized neighborhoods. FT No-Graph is fine-tuned without neighborhood inputs. Bold and underline indicate the best and second-best results.}
\label{tab:vlm_controls}
\end{table}
As shown in Table~\ref{tab:vlm_controls}, OMG-VLM outperforms the matched fine-tuned no-graph control on all nine datasets, with an average improvement of \textbf{12.62 points}. Because this control retains the same pretrained backbone and task-adaptation setting, the improvement demonstrates that graph neighborhoods and their learned aggregation provide consistent performance gains beyond target-node attributes and fine-tuning alone.

OMG-VLM also consistently outperforms both zero-shot controls. The gaps are particularly large on arXiv$^{\text{\scriptsize NC}}$ (\textbf{67.85} vs.\ 8.50), Movies$^{\text{\scriptsize LP}}$ (\textbf{95.30} vs.\ 61.15), RedditS$^{\text{\scriptsize LP}}$ (\textbf{99.10} vs.\ 62.68), and RedditM$^{\text{\scriptsize LP}}$ (\textbf{92.45} vs.\ 68.60). Structural prompting does not consistently improve over non-structural prompting. These results show that directly serializing graph context into a pretrained VLM is insufficient and further support the value of trained graph adapters for aggregating neighborhood information within the VLM's native representation space.
\subsection{Layer $L$ Sensitivity Analysis}
\label{app:layer_sensitivity}

\begin{figure}[t]
\centering
\vspace{-0.8em}
\resizebox{0.72\linewidth}{!}{
\begin{tikzpicture}
\begin{axis}[
    width=0.98\linewidth,
    height=5cm,
    ybar,
    bar width=18pt,
    ymin=90.5,
    ymax=91.7,
    ylabel={Accuracy (\%)},
    symbolic x coords={$L=1$,$L=2$,$L=3$},
    xtick=data,
    xticklabel style={font=\scriptsize},
    yticklabel style={font=\scriptsize},
    ylabel style={font=\scriptsize},
    ymajorgrids=true,
    grid style={dashed,gray!25},
    axis y line*=left,
    axis x line*=bottom,
    tick align=outside,
    axis line style={black!70},
    enlarge x limits=0.35,
    nodes near coords,
    every node near coord/.append style={
        font=\scriptsize,
        anchor=south
    },
]

\addplot[
    fill=orange!35,
    draw=orange!85!black,
    line width=0.8pt,
] coordinates {
    ($L=1$,91.45)
    ($L=2$,90.94)
    ($L=3$,91.20)
};

\end{axis}
\end{tikzpicture}
}
\vspace{-0.7em}
\caption{Sensitivity to the number of visual aggregation layers $L$. Results are reported on Arts$^{\text{\scriptsize NC}}$ with separate models trained and evaluated for each value of $L$.}
\label{fig:layer_sensitivity}
\vspace{-0.8em}
\end{figure}
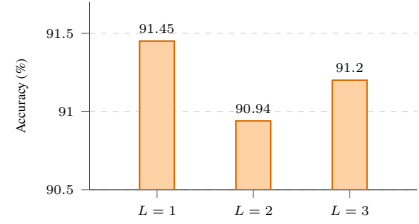

We analyze the sensitivity of OMG-VLM to the number of center-conditioned visual aggregation layers $L$ in the Graph-Aware Visual Adapter. For efficiency, we conduct a targeted sensitivity analysis on Arts$^{\text{\scriptsize NC}}$, training and evaluating separate models with $L \in \{1,2,3\}$, rather than repeating the full mixed-training protocol for each setting. Figure~\ref{fig:layer_sensitivity} reports the results.

The results suggest that performance is stable across the tested values of $L$. All variants remain within a narrow range, with the best result obtained at $L=1$. Since deeper aggregation does not yield consistent improvement in this analysis, we use a small value of $L=1$ in the main model to reduce unnecessary computation.

\section{Prompt Templates and Qualitative Examples}
\label{app:prompts_examples}

\subsection{Prompt Templates}
\label{app:prompt_template}

\newcommand{\tok}[1]{\textcolor{blue}{\texttt{<#1>}}}

Table~\ref{tab:prompt_templates_nc} and Table~\ref{tab:prompt_templates_lp} summarize the prompt templates used for node classification and link prediction, respectively.
All prompts follow the ChatML format, where the system message is set to ``You are a helpful assistant.''~\cite{bai_qwen-vl_2023}.
The placeholders \tok{anchor text} and \tok{anchor image} denote the textual and visual attributes of the anchor node, while \tok{text neighborhood} and \tok{image neighborhood} denote modality-specific neighborhood context incorporated into the input.

\begin{table*}[t]
\centering
\small
\setlength{\tabcolsep}{5pt}
\renewcommand{\arraystretch}{1.15}
\begin{tabular}{p{0.12\linewidth}p{0.08\linewidth}p{0.72\linewidth}}
\toprule
\textbf{Dataset} & \textbf{Type} & \textbf{User Prompt Content} \\
\midrule

Arts & MAG &
Given the target product information on Amazon:
Picture: \tok{anchor image}
Image neighborhood: \tok{image neighborhood}
Title and description: \tok{anchor text}
Text neighborhood: \tok{text neighborhood}
Question: Based on the target product's picture, title, description, and neighborhood context, which category does the target product belong to? Choose from the following options:
Knitting \& Crochet; Beading \& Jewelry Making; Painting, Drawing \& Art Supplies; Crafting; Model \& Hobby Building; Sewing; Scrapbooking \& Stamping. \\

\midrule

arXiv & TAG &
Given the target paper information on arXiv:
Title and abstract: \tok{anchor text}
Text neighborhood: \tok{text neighborhood}
Question: Based on the target paper's title, abstract, and neighborhood context, which of the following categories does the target paper belong to?
Options: arxiv cs ai, arxiv cs ar, arxiv cs cc, arxiv cs ce, arxiv cs cg, arxiv cs cl, arxiv cs cr, arxiv cs cv, arxiv cs cy, arxiv cs db, arxiv cs dc, arxiv cs dl, arxiv cs dm, arxiv cs ds, arxiv cs et, arxiv cs fl, arxiv cs gl, arxiv cs gr, arxiv cs gt, arxiv cs hc, arxiv cs ir, arxiv cs it, arxiv cs lg, arxiv cs lo, arxiv cs ma, arxiv cs mm, arxiv cs ms, arxiv cs na, arxiv cs ne, arxiv cs ni, arxiv cs oh, arxiv cs os, arxiv cs pf, arxiv cs pl, arxiv cs ro, arxiv cs sc, arxiv cs sd, arxiv cs se, arxiv cs si, arxiv cs sy. \\

\midrule

PubMed & TAG &
The following is the title and the abstract of a paper:
Title and abstract: \tok{anchor text}
Text neighborhood: \tok{text neighborhood}
Question: Based on the paper's title, abstract, and neighborhood context, which case of diabetes does the paper involve? Choose from the following options:
Type 1 diabetes; Type 2 diabetes; Experimentally induced diabetes. \\

\midrule

CDs & MAG &
Given the target CD/Vinyl product information on Amazon:
Picture: \tok{anchor image}
Image neighborhood: \tok{image neighborhood}
Title and description: \tok{anchor text}
Text neighborhood: \tok{text neighborhood}
Question: Based on the target product's picture, title, description, and neighborhood context, which category does the target product belong to? Choose from the following options:
Pop; Today's Deals in Music; Rock; Indie \& Alternative; Classic Rock; Country; International Music; Jazz; Metal; R\&B; Classical; Rap \& Hip-Hop; Christian \& Gospel; Blues; Dance \& Electronic. \\

\bottomrule
\end{tabular}
\caption{Prompt templates for node classification datasets. Each prompt explicitly separates anchor-node attributes from modality-specific neighborhood context.}
\label{tab:prompt_templates_nc}
\end{table*}

\begin{table*}[t]
\centering
\small
\setlength{\tabcolsep}{5pt}
\renewcommand{\arraystretch}{1.15}
\begin{tabular}{p{0.12\linewidth}p{0.08\linewidth}p{0.72\linewidth}}
\toprule
\textbf{Dataset} & \textbf{Type} & \textbf{User Prompt Content} \\
\midrule

RedditS & IAG &
Given two nodes from a social media graph, the information of the first node is as follows:
Picture: \tok{anchor image}
Image neighborhood: \tok{image neighborhood}
and the information of the other node is:
Picture: \tok{anchor image}
Image neighborhood: \tok{image neighborhood}
If the connections between nodes represent the co-comment relationships between posts, are these two central nodes connected? Give me a direct answer of ``yes'' or ``no''. \\

\midrule

Movies & MAG &
Given two nodes from the Amazon movies product graph, the information of the first node is as follows:
Picture: \tok{anchor image};
Image neighborhood: \tok{image neighborhood};
Title and description: \tok{anchor text}
Text neighborhood: \tok{text neighborhood}
and the information of the other node is:
Picture: \tok{anchor image};
Image neighborhood: \tok{image neighborhood};
Title and description: \tok{anchor text}
Text neighborhood: \tok{text neighborhood}
If the connections between nodes represent the co-purchased or co-reviewed relationships between products, are these two central nodes connected? Give me a direct answer of ``yes'' or ``no''. \\

\midrule

Cora & TAG &
Given two nodes from the Cora citation graph of machine learning papers, the information of the first node is as follows:
Title and abstract: \tok{anchor text}
Text neighborhood: \tok{text neighborhood}
And the information of the other node is:
Title and abstract: \tok{anchor text}
Text neighborhood: \tok{text neighborhood}
If the connections between nodes represent citation relationships, i.e., one paper cites the other as a reference in its work, are these two central nodes connected? Give me a direct answer of ``yes'' or ``no''. \\

\midrule

RedditM & MAG &
Given two nodes from a social media graph, the information of the first node is as follows:
Picture: \tok{anchor image};
Image neighborhood: \tok{image neighborhood};
Text content: \tok{anchor text}
Text neighborhood: \tok{text neighborhood}
and the information of the other node is:
Picture: \tok{anchor image};
Image neighborhood: \tok{image neighborhood};
Text content: \tok{anchor text}
Text neighborhood: \tok{text neighborhood}
If the connections between nodes represent the co-comment relationships between posts, are these two central nodes connected? Give me a direct answer of ``yes'' or ``no''. \\

\midrule

VideoGames & MAG &
Given two nodes from the Amazon video games product graph, the information of the first node is as follows:
Picture: \tok{anchor image};
Image neighborhood: \tok{image neighborhood};
Title and description: \tok{anchor text}
Text neighborhood: \tok{text neighborhood}
and the information of the other node is:
Picture: \tok{anchor image};
Image neighborhood: \tok{image neighborhood};
Title and description: \tok{anchor text}
Text neighborhood: \tok{text neighborhood}
If the connections between nodes represent the co-purchased or co-reviewed relationships between products, are these two central nodes connected? Give me a direct answer of ``yes'' or ``no''. \\

\bottomrule
\end{tabular}
\caption{Prompt templates for link prediction datasets. Each prompt presents the two anchor nodes together with their modality-specific neighborhood context.}
\label{tab:prompt_templates_lp}
\end{table*}

% =========================
% Qualitative Examples
% =========================

\subsection{Qualitative Examples}
\label{app:qualitative_examples}

We show three qualitative examples from different datasets, each including the constructed input, retrieved graph context, model response, and ground-truth answer.

\begin{figure*}[t]
\centering
\begin{casebox}{MAG Amazon-Arts: Node Classification}

\casefield{Target Product}
\begin{minipage}[t]{0.26\textwidth}
\centering
\includegraphics[width=2.6cm]{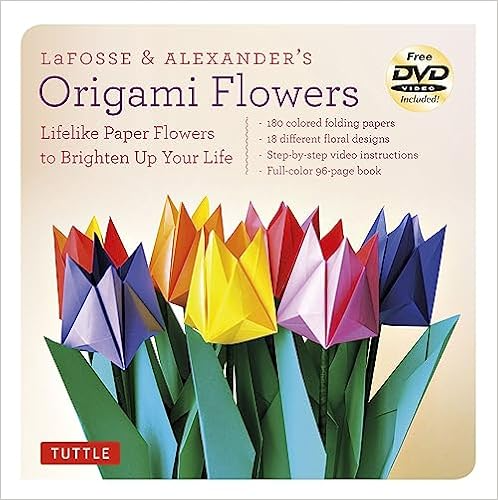}
\end{minipage}
\hfill
\begin{minipage}[t]{0.58\textwidth}
\casetext{\textbf{Title and Description.}
\textit{Origami Paper - Traditional Prints - 8 1/4'' - 49 Sheets: Tuttle Origami Paper: Large Origami Sheets Printed with 6 Different Patterns: Instructions for 6 Projects Included. From the Back Cover About the Author Tuttle Studio draws......}}
\end{minipage}

\vspace{5pt}
\casefield{Retrieved Neighborhood Context}
\begin{center}
\begin{tabular}{>{\centering\arraybackslash}p{0.30\textwidth}
                >{\centering\arraybackslash}p{0.30\textwidth}
                p{0.30\textwidth}}
\includegraphics[width=2.2cm]{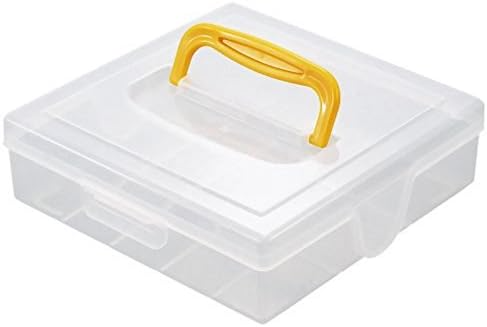} &
\includegraphics[width=2.2cm]{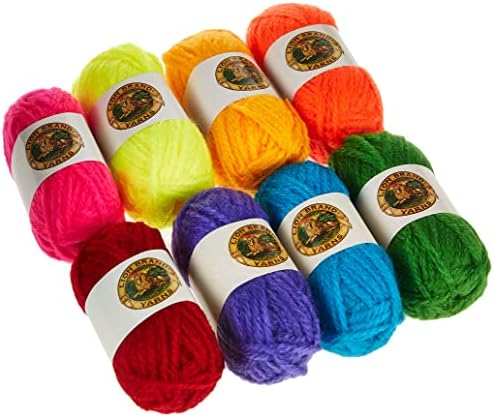} &
\footnotesize\textit{[Additional neighbors omitted for brevity]} \\
\footnotesize Neighbor 1: \textit{LuPro Japanese LPAT-4006 Origami Folding Paper Case Box......} &
\footnotesize Neighbor 2: \textit{Lion Brand Yarn Lion Brand Bonbons 680 Crayons......} &
\\
\end{tabular}
\end{center}

\casefield{Prompt}
\begin{tcolorbox}[
  colback=white,
  colframe=gray!25,
  boxrule=0.4pt,
  arc=1pt,
  left=4pt,
  right=4pt,
  top=3pt,
  bottom=3pt
]
\footnotesize
Given the target product information on Amazon:

Picture: \promptbox{<anchor image>}

Title and description: \promptbox{<anchor text>}

The target product is associated with the following graph neighborhood context:

Neighbor pictures: \promptbox{<image neighborhood>}

Neighbor titles and descriptions: \promptbox{<text neighborhood>}

Question: Based on the target product's picture, title, description, and graph neighborhood context, which category does the target product belong to?

Options: Knitting \& Crochet, Beading \& Jewelry Making, Painting, Drawing \& Art Supplies, Crafting, Model \& Hobby Building, Sewing, Scrapbooking \& Stamping.
\end{tcolorbox}

\answerrow{Crafting}{Crafting}

\end{casebox}
\caption{Qualitative example for MAG Amazon-Arts node classification. Full text information is omitted for brevity.}
\label{fig:prompt_arts}
\end{figure*}

\noindent\textbf{Amazon-Arts.}
Figure~\ref{fig:prompt_arts} shows a MAG node classification example where OMG-VLM predicts the product category from target attributes and graph-aware multimodal neighborhood context.

\noindent\textbf{ogbn-arXiv.}
Figure~\ref{fig:prompt_arxiv} shows a TAG node classification example where OMG-VLM predicts the paper category from textual attributes and graph-aware textual neighborhood context.

\begin{figure*}[t]
\centering
\begin{casebox}{IAG RedditS: Link Prediction}

\casefield{Target Node Pair}
\begin{center}
\begin{tabular}{>{\centering\arraybackslash}p{0.42\textwidth}
                >{\centering\arraybackslash}p{0.42\textwidth}}
\textbf{\small Node A} & \textbf{\small Node B} \\
\includegraphics[width=3.4cm]{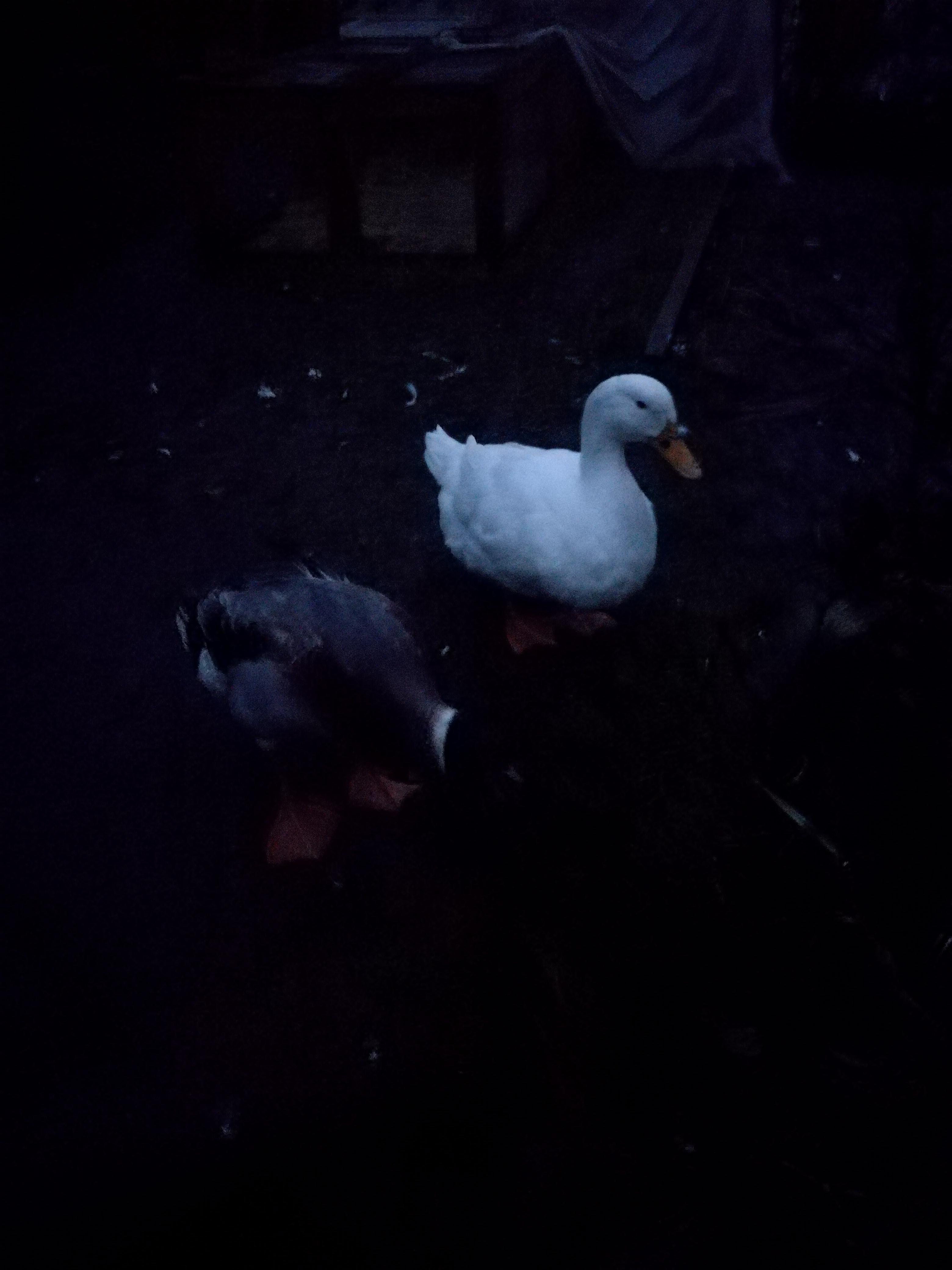} &
\includegraphics[width=3.4cm]{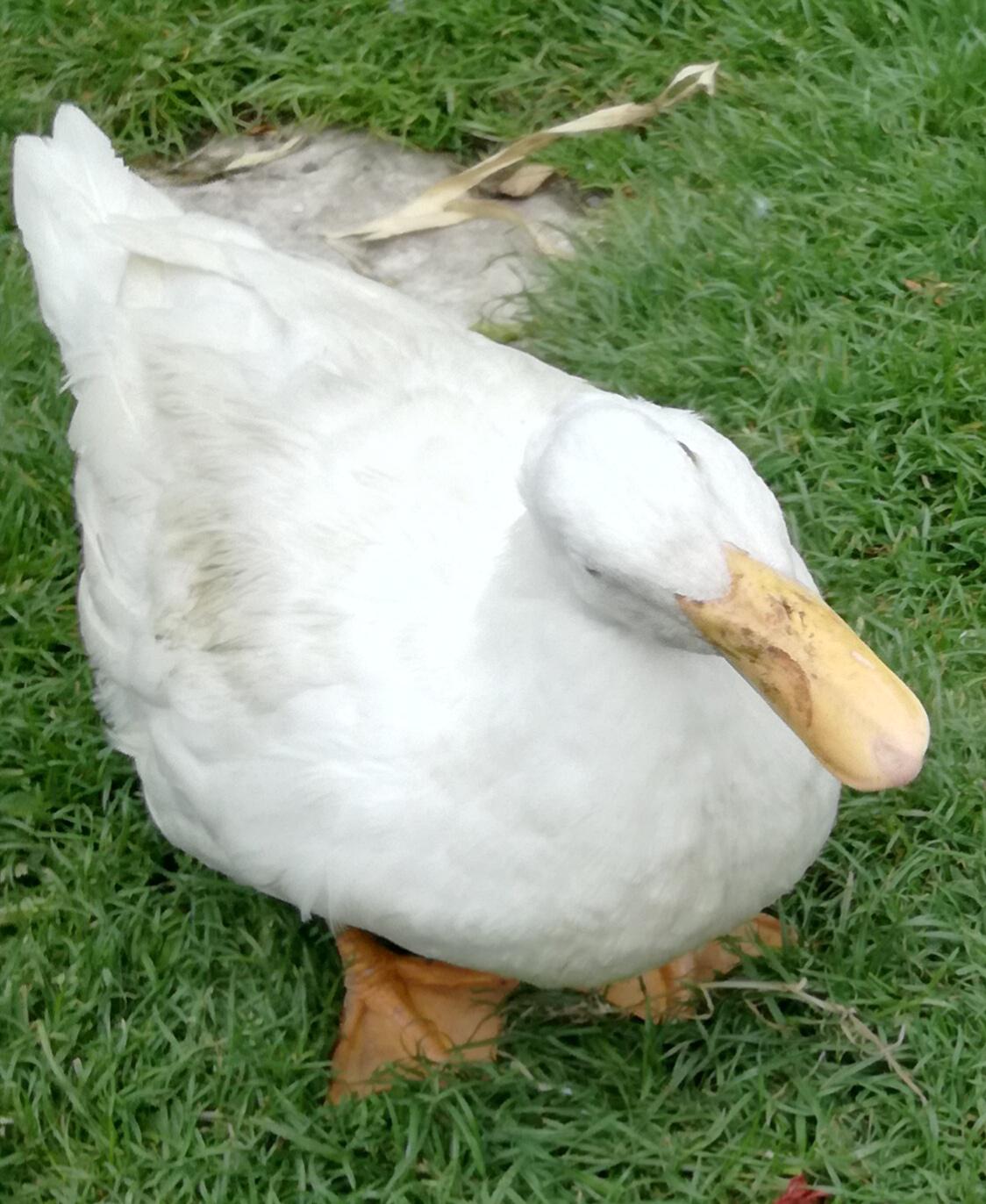} \\
\end{tabular}
\end{center}

\casefield{Retrieved Neighborhood Context}
\begin{center}
\begin{tabular}{>{\centering\arraybackslash}p{0.42\textwidth}
                >{\centering\arraybackslash}p{0.42\textwidth}}
\textbf{\small Neighbor of Node A} & \textbf{\small Neighbor of Node B} \\
\includegraphics[width=3.0cm]{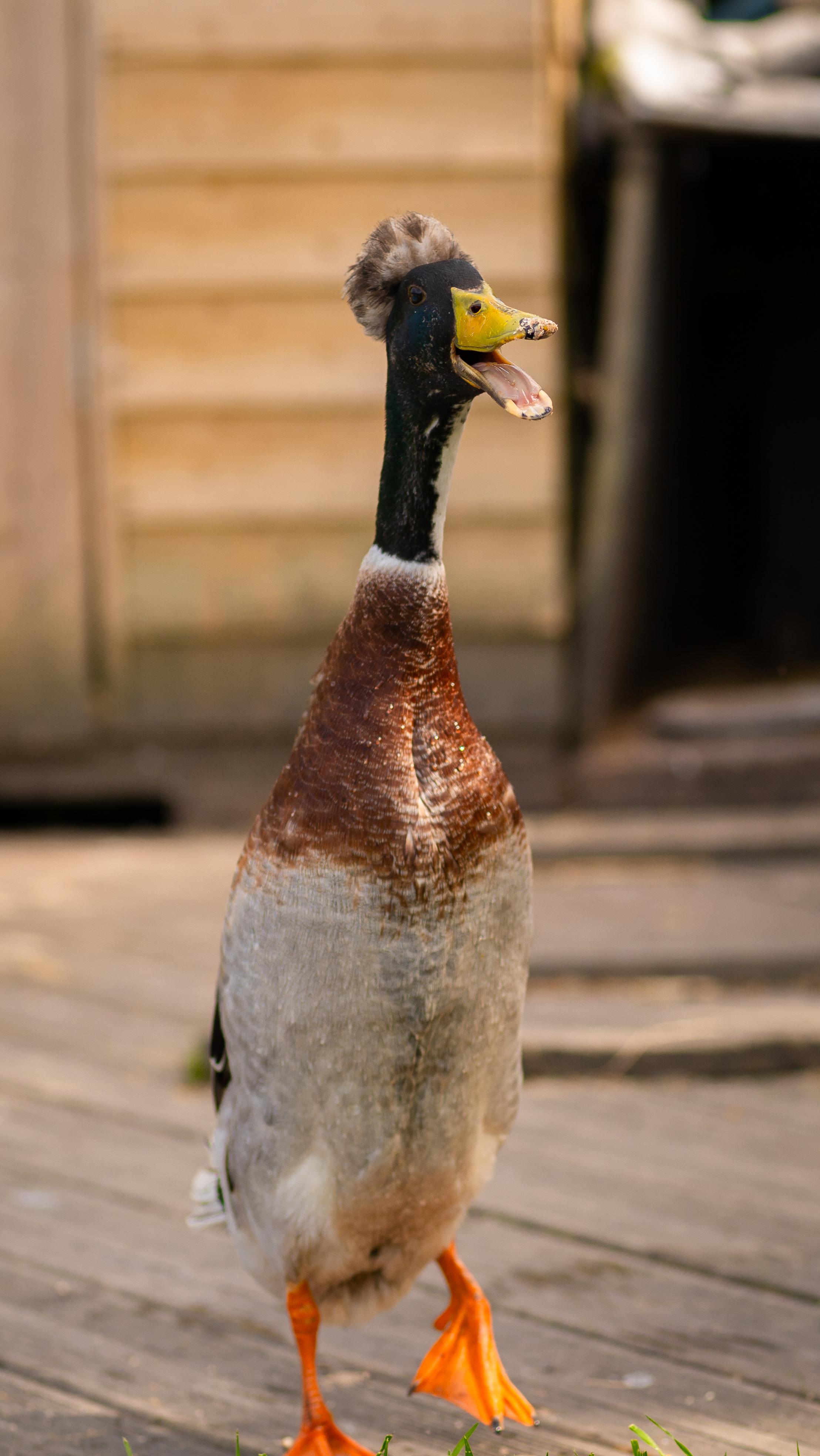} &
\includegraphics[width=3.0cm]{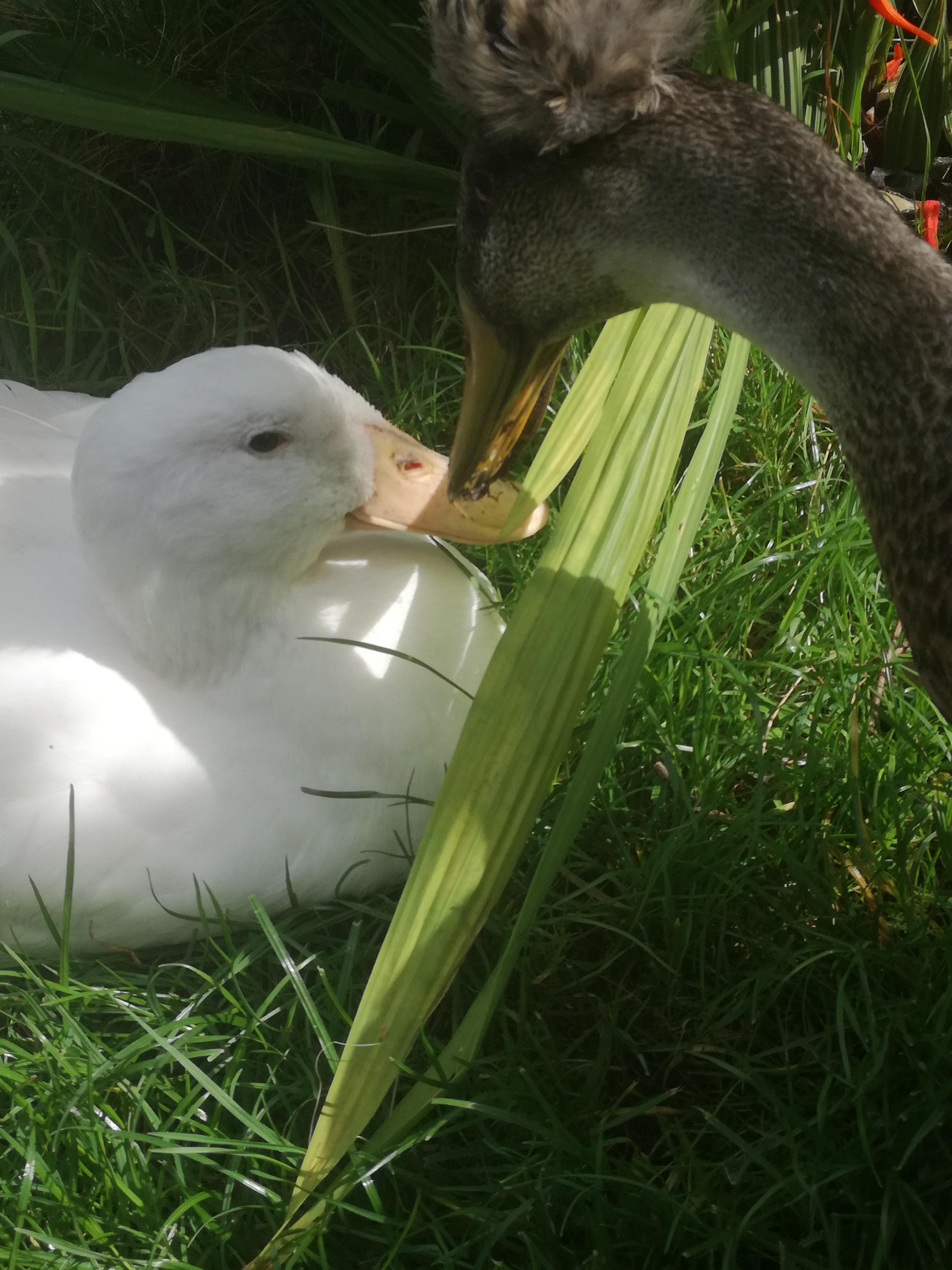} \\
\end{tabular}
\end{center}

\vspace{-2pt}
\casetext{\textit{[Additional neighbors omitted for brevity]}}

\casefield{Prompt}
\begin{tcolorbox}[
  colback=white,
  colframe=gray!25,
  boxrule=0.4pt,
  arc=1pt,
  left=4pt,
  right=4pt,
  top=3pt,
  bottom=3pt
]
\footnotesize
Given two nodes from a social media graph, the information of the first node is as follows:

Picture: \promptbox{<anchor image of Node A>}

The first node is associated with the following graph neighborhood context:

Neighbor pictures: \promptbox{<image neighborhood of Node A>}

The information of the second node is as follows:

Picture: \promptbox{<anchor image of Node B>}

The second node is associated with the following graph neighborhood context:

Neighbor pictures: \promptbox{<image neighborhood of Node B>}

If the connections between nodes represent the co-comment relationships between posts, are these two central nodes connected? Give a direct answer of ``yes'' or ``no''.
\end{tcolorbox}

\answerrow{yes}{yes}

\end{casebox}
\caption{Qualitative example for IAG RedditS link prediction.}
\label{fig:prompt_reddits}
\end{figure*}

\noindent\textbf{RedditS.}
Figure~\ref{fig:prompt_reddits} shows an IAG link prediction example where OMG-VLM predicts whether two image-attributed posts are connected using their visual attributes and neighborhood context.

\begin{figure*}[t]
\centering
\begin{casebox}{TAG ogbn-arXiv: Node Classification}

\casefield{Target Paper}
\casetext{\textbf{Title and Abstract.}
\textit{learning spectral spatial temporal features via a recurrent convolutional neural network for change detection in multispectral imagery Change detection is one of the central problems......}}

\vspace{5pt}
\casefield{Retrieved Neighborhood Context}
\begin{tcolorbox}[
  colback=white,
  colframe=gray!25,
  boxrule=0.4pt,
  arc=1pt,
  left=4pt,
  right=4pt,
  top=3pt,
  bottom=3pt
]
\footnotesize
\textbf{Neighbor 1.} \textit{multitask learning for large scale semantic change detection Change detection is one of the main problems in remote sensing, and is essential to the accurate processing and understanding......}

\vspace{2pt}
\textbf{Neighbor 2.} \textit{deeplab semantic image segmentation with deep convolutional nets atrous convolution and fully connected crfs In this work we address the task of semantic image segmentation with Deep Learning......}

\vspace{2pt}
\textit{[Additional neighbors omitted for brevity]}
\end{tcolorbox}

\casefield{Prompt}
\begin{tcolorbox}[
  colback=white,
  colframe=gray!25,
  boxrule=0.4pt,
  arc=1pt,
  left=4pt,
  right=4pt,
  top=3pt,
  bottom=3pt
]
\footnotesize
Given the target paper information on arXiv:

Title and abstract: \promptbox{<anchor text>}

The target paper is associated with the following graph neighborhood context:

Neighbor titles and abstracts: \promptbox{<text neighborhood>}

Question: Based on the target paper's title, abstract, and graph neighborhood context, which of the following categories does the target paper belong to?

Options: arxiv cs ai, arxiv cs ar, arxiv cs cc, arxiv cs ce, arxiv cs cg, arxiv cs cl, arxiv cs cr, arxiv cs cv, arxiv cs cy, arxiv cs db, arxiv cs dc, arxiv cs dl, arxiv cs dm, arxiv cs ds, arxiv cs et, arxiv cs fl, arxiv cs gl, arxiv cs gr, arxiv cs gt, arxiv cs hc, arxiv cs ir, arxiv cs it, arxiv cs lg, arxiv cs lo, arxiv cs ma, arxiv cs mm, arxiv cs ms, arxiv cs na, arxiv cs ne, arxiv cs ni, arxiv cs oh, arxiv cs os, arxiv cs pf, arxiv cs pl, arxiv cs ro, arxiv cs sc, arxiv cs sd, arxiv cs se, arxiv cs si, arxiv cs sy.
\end{tcolorbox}

\answerrow{arxiv cs cv}{arxiv cs cv}

\end{casebox}
\caption{Qualitative example for TAG ogbn-arXiv node classification. Full text information is omitted for brevity.}
\label{fig:prompt_arxiv}
\end{figure*}

\end{document}